\documentclass[conference]{IEEEtran}
\IEEEoverridecommandlockouts
\usepackage{cite}
\usepackage{amsmath,amssymb,amsfonts}
\usepackage{graphicx}
\usepackage{textcomp}
\usepackage{xcolor}
\def\BibTeX{{\rm B\kern-.05em{\sc i\kern-.025em b}\kern-.08em
    T\kern-.1667em\lower.7ex\hbox{E}\kern-.125emX}}

\usepackage{eurosym}
\usepackage{array}
\usepackage{subfigure}

\usepackage{booktabs}
\usepackage{multirow}
\usepackage{caption}
\usepackage{dblfloatfix}

\usepackage[noend]{algorithm2e,algorithmic}
\usepackage{float}
\usepackage{psfrag}
\usepackage{hyperref}
\hypersetup{
    colorlinks,
    citecolor=blue,
    filecolor=blue,
    linkcolor=blue,
    urlcolor=blue,
    pdfstartview={XYZ null null 1.00}
}

\definecolor{gray1}{gray}{0.7}
\definecolor{gray2}{gray}{0.9}

\begin{document}

\title{Multi-criteria Evolution of Neural Network Topologies: Balancing Experience and Performance in Autonomous Systems
\thanks{Copyright\copyright 2018 ASME. Personal use of this material is permitted. Permission from ASME must be obtained for all other uses, in any current or future media, including reprinting/republishing this material for advertising or promotional purposes, creating new collective works, for resale or redistribution to servers or lists, or reuse of any copyrighted component of this work in other works.}
}

\author{\IEEEauthorblockN{Sharat Chidambaran\IEEEauthorrefmark{1},
Amir Behjat\IEEEauthorrefmark{2}, and
Souma Chowdhury\IEEEauthorrefmark{3}}
\IEEEauthorblockA{\textit{Department of Mechanical \& Aerospace Engineering} \\
\textit{University at Buffalo}\\
Buffalo, NY, 14260\\
Email: \IEEEauthorrefmark{1}sharatpa@buffalo.edu,
\IEEEauthorrefmark{2}amirbehj@buffalo.edu,
\IEEEauthorrefmark{3}soumacho@buffalo.edu}}

% \author{\IEEEauthorblockN{Sharat Chidambaran }
% \IEEEauthorblockA{\textit{Department of Mechanical \& Aerospace Engineering} \\
% \textit{University at Buffalo}\\
% Buffalo, NY, 14260\\
% sharatpa@buffalo.edu}
% \and
% \IEEEauthorblockN{Amir Behjat}
% \IEEEauthorblockA{\textit{Department of Mechanical \& Aerospace Engineering} \\
% \textit{University at Buffalo}\\
% Buffalo, NY, 14260\\
% amirbehj@buffalo.edu}
% \and
% \IEEEauthorblockN{Souma Chowdhury}
% \IEEEauthorblockA{\textit{Department of Mechanical \& Aerospace Engineering} \\
% \textit{University at Buffalo}\\
% Buffalo, NY, 14260\\
% soumacho@buffalo.edu}
% }

\maketitle
%%%%%%%%%%%%%%%%%%%%%%%%%%%%%%%%%%%%%%%%%%%%%%%%%%%%%%%%%%%%%%%%%%%%%%
\begin{abstract}Majority of Artificial Neural Network (ANN) implementations in autonomous systems use a fixed/user-prescribed network topology, leading to sub-optimal performance and low portability. The existing neuro-evolution of augmenting topology or NEAT paradigm offers a powerful alternative by allowing the network topology and the connection weights to be simultaneously optimized through an evolutionary process. However, most NEAT implementations allow the consideration of only a single objective. There also persists the question of how to tractably introduce topological diversification that mitigates overfitting to training scenarios. To address these gaps, this paper develops a multi-objective neuro-evolution algorithm. While adopting the basic elements of NEAT, important modifications are made to the selection, speciation, and mutation processes. With the backdrop of small-robot path-planning applications, an experience-gain criterion is derived to encapsulate the amount of diverse local environment encountered by the system. This criterion facilitates the evolution of genes that support exploration, thereby seeking to generalize from a smaller set of mission scenarios than possible with performance maximization alone. The effectiveness of the single-objective (optimizing performance) and the multi-objective (optimizing performance and experience-gain) neuro-evolution approaches are evaluated on two different small-robot cases, with ANNs obtained by the multi-objective optimization observed to provide superior performance in unseen scenarios.
\end{abstract}

%\vspace{-1.2cm}

%%%%%%%%%%%%%%%%%%%%%%%%%%%%%%%%%%%%%%%%%%%%%%%%%%%%%%%%%%%%%%%%%%%%%%
\section{INTRODUCTION}
\subsection{Neural Network as Decision-Support}
%Autonomous systems must be able to operate in a variety of scenarios, without the need of human intervention, and with an acceptable level of performance. Traditionally, controllers for autonomous systems, such as robots, employed elaborate mathematical models - this gave way to the introduction of fuzzy logic controllers~\cite{saffiotti1997uses}. These controllers, however, often produced undesirable behaviors and were incapable of performing any sort of learning. This provided an impetus for researchers to look for an alternative, heuristic-based controllers - a very popular and powerful class of them being neural networks.
Artificial Neural networks or ANNs are becoming a popular model for higher level decision-support or planning in various intelligent systems\cite{dounis2009advanced,lewis1998neural}.
%[REF -- \textcolor{red}{give examples from robotics, energy/smart-grid, IoT}].
This emergence is partly attributed to the capability of ANNs to serve as universal function approximators, both for discrete (classification) and continuous prediction (regression) \cite{hagan1996neural}. In this paper, we focus on the construction and usage of ANNs to serve as a planner in small autonomous vehicular systems (particularly small ground robots) \cite{yang2000efficient,duan2014imperialist}. In such autonomous system applications, the architecture or topology of the ANN is often user-prescribed, thereby leading to sub-optimal prediction models \cite{turner2013importance} that perform state-to-action mapping.
%[REF -- \textcolor{red}{just cite recent NEAT papers that have mentioned these drawbacks of fixed topology NN}].
%
\begin{figure*}
\vspace{-0.5cm}
\centering
\begin{tabular}{cc}
\includegraphics[width =1.0 \columnwidth]{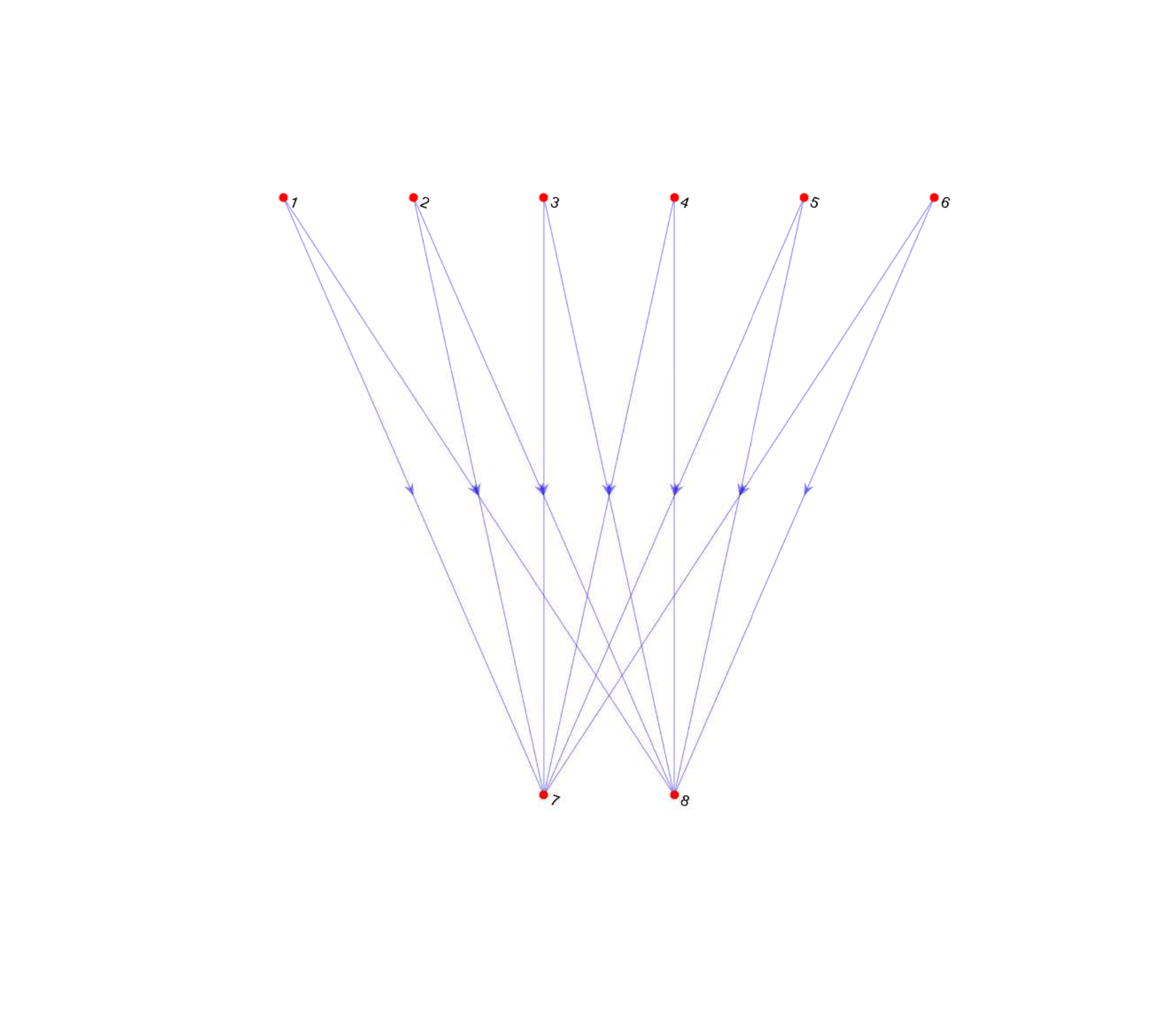} &
\includegraphics[width =1.0\columnwidth]{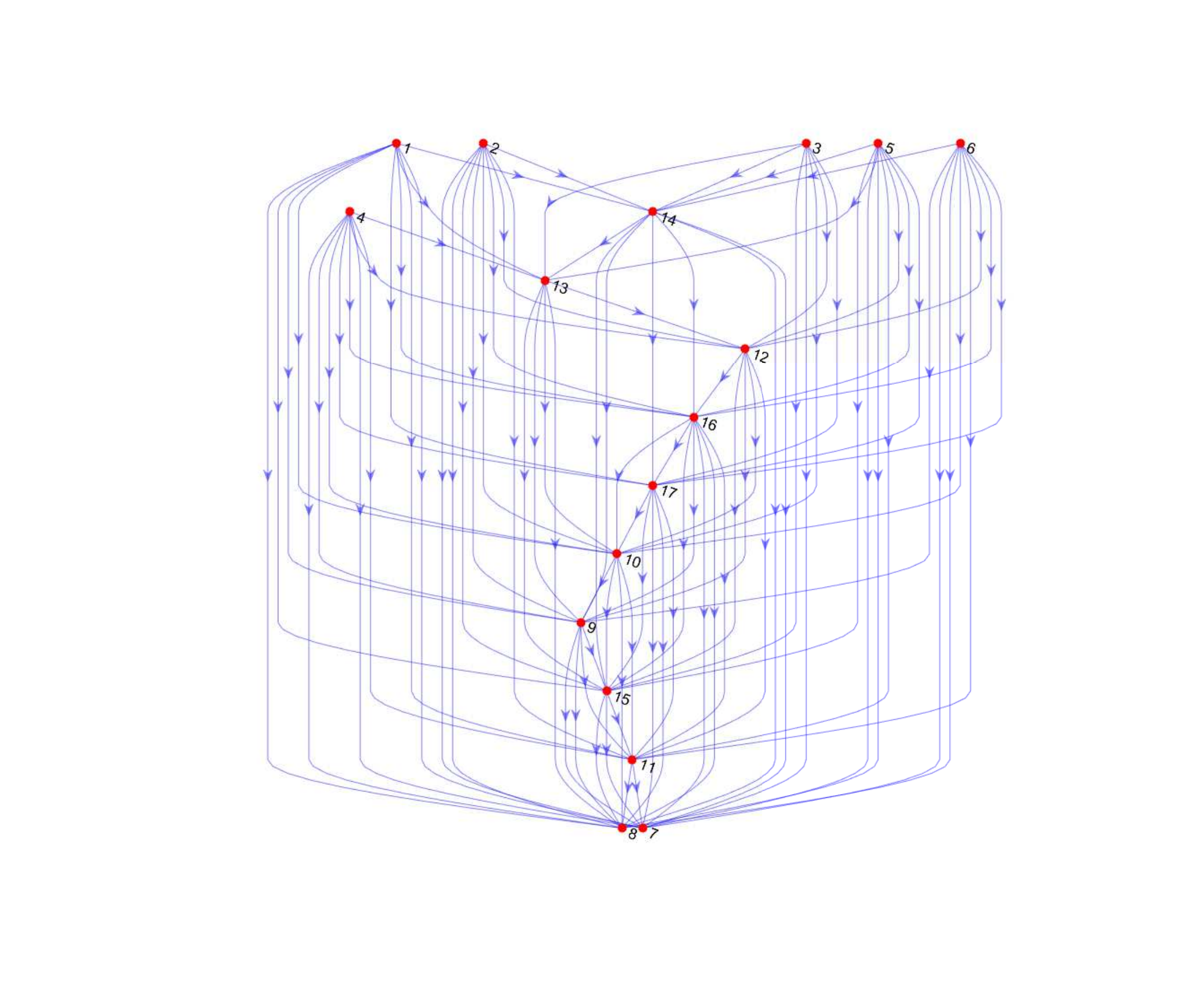}\vspace{-1cm}\\
(a) Initial Neural Network Topology & (b) Evolved Neural Network Topology
\end{tabular}
\captionsetup{justification=centering}
\caption{Variation in Network Topologies via Neuroevolution}
\label{ini_network}
\vspace{-0.5 cm}
\end{figure*}

\begin{figure*}
\vspace{-0.5cm}
\centering
\begin{tabular}{c c}
\includegraphics[page=1,width = 1\columnwidth]{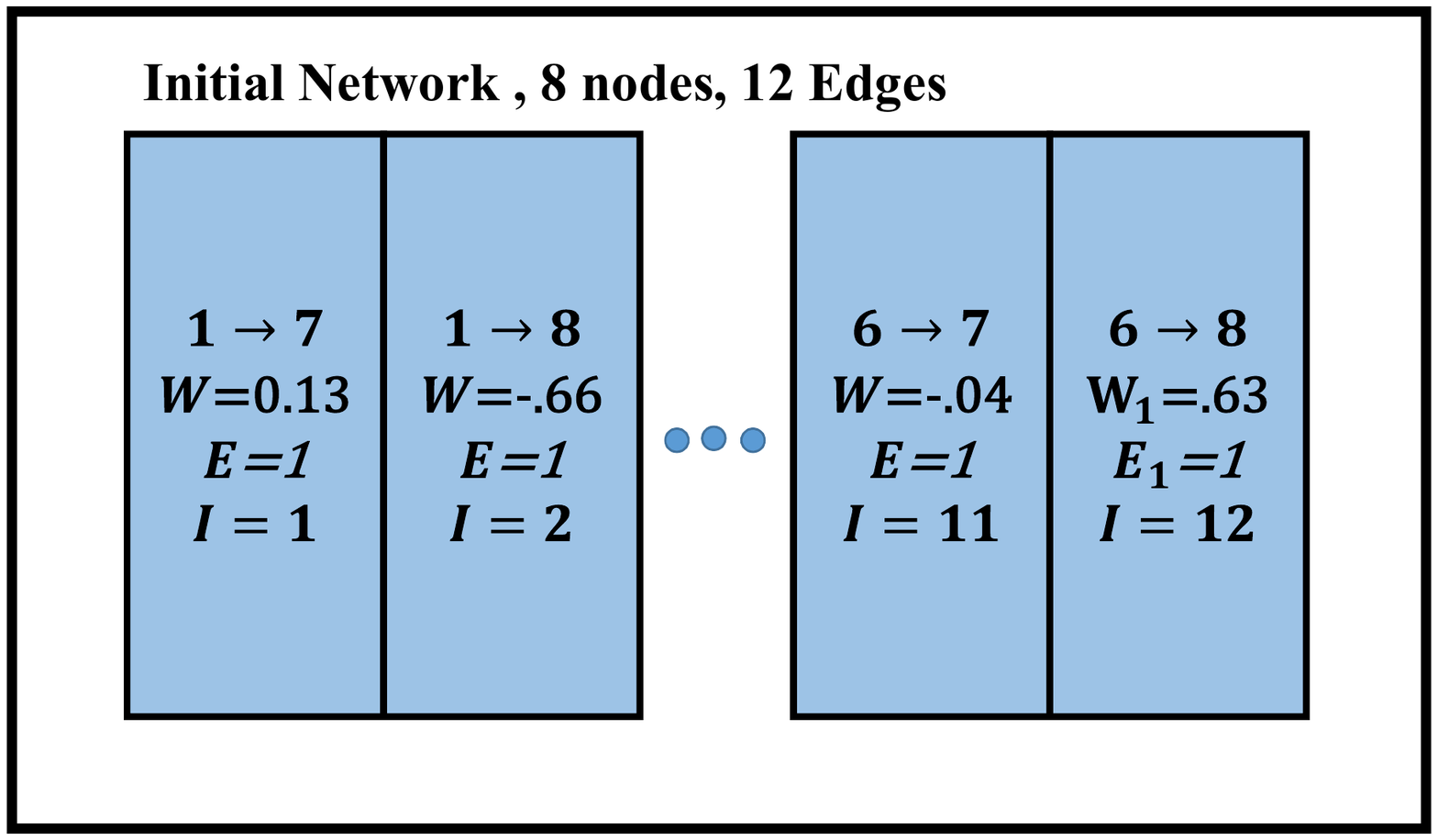} &
\includegraphics[page=2,width =1\columnwidth]{netw.pdf}\vspace{-1.5 cm}
\\

(a) Genetic Encoding of the Initial Network & (b) Genetic Encoding of the Evolved Network\\
\end{tabular}
\captionsetup{justification=centering}
\caption{Genetic Encoding of the Neural Networks}
\label{final_network}
\end{figure*}
% \begin{figure*}
% \centering
% \subfigure[Neural Network Topology at the Start of the Optimization Process]{%
% \label{simplearch}%
% \includegraphics[width=1.2\columnwidth]{iniitneat.pdf}}%
% \qquad
% \subfigure[Neural Network Topology at the End of the Optimization Process]{%
% \label{complexarch}%
% \includegraphics[width=1.0\columnwidth]{1_stage_final.pdf}}%
% \captionsetup{justification=centering}
% \caption{Complexification of Neural Network Topology through neuro-evolution} \label{fig:NEAT}
% \end{figure*}
%

Additional barriers to effective use of ANNs in autonomous systems decision-support are presented by the lack/scarcity of prior data for labeling, or the prohibitive expense in collecting/generating it at a research lab or small business scale \cite{forman2004learning}. The ``evolution of neural network topologies" paradigm\cite{maniezzo1994genetic} that has mostly emerged in the past 20 years offers exciting opportunities to tackle these challenges. This paper seeks to make essential contributions to this paradigm by developing a \textit{multi-criteria} approach to evolving neural topologies and exploring the capacity to enable meaningful decision-making novelty through an experience gain concept. Preliminary evaluation of these new contributions to neuro-evolution is performed through application to small unmanned ground vehicles (UGV) planning problems.

In this section, we provide a brief survey of literature in ANN-based decision-support implementations and their associated challenges, followed by a review of neuroevolution methods and novelty search. %(surveying the literature in the broad area of path-planning in autonomous systems is however not within the scope of this paper).
Back in 1990, Pomerleau\cite{pomerleau1990neural} demonstrated that a neural network with a single hidden layer could perform a road-following task despite the presence of noise in sensor inputs and environmental uncertainty. Later implementations of ANNs catered to generating more complex behaviors (in reasoning, adaptation, and learning) -- e.g., a hybrid intelligent system combining ANNs, a genetic algorithm, and fuzzy logic was developed by the NASA ACE center \cite{akbarzadeh2000soft} to produce wall-following, obstacle avoidance, and goal-seeking capabilities in robots. More recent implementations have explored complex learning formalisms such as deep ANN training via expert demonstrations for motion planning \cite{pfeiffer2017perception}. Other recent notable examples of the use of ANN in the domain of autonomous systems can be found in \cite{wu2016squeezedet,park2015neural}.

%\vspace{-0.5cm}
\subsection{Neuro-Evolution}
To a large extent, current implementations of ANNs have employed fixed-topology neural networks in which only the weights are optimized via training methods such as backpropagation. Neuro-evolution is the process of employing evolutionary algorithms to evolve neural networks, which can also work in the absence of labeling (i.e., as an unsupervised approach to optimal decision-support training). By simultaneously evolving the topology and connection weights through a genetic algorithm-type approach (but with the specialized direct encoding of genotypes and crossover and mutation operators), Stanley et al. \cite{stanley2002efficient} demonstrated that it is possible to outperform some of the best fixed topology networks in benchmark reinforcement learning problems. This approach is known as Neuro-evolution of Augmenting Topologies or NEAT. Although the ``Neuro-evolution" paradigm\cite{ronald1994genetic} predates NEAT, the latter pioneered a new direction in neuro-evolution that enabled: i) adaptively-incremental growth in (as opposed to randomized exploration of) topological complexity and ii) preservation of topological innovation during the evolutionary search process\cite{stanley2002evolving,stanley2009hypercube,vargas2017spectrum} %[REF -- \textcolor{red}{cite multiple main papers on NEAT and HyperNEAT}].
For the sake of providing a quick visual introduction to what NEAT accomplishes, we respectively provide, in Fig. \ref{ini_network} and Fig. \ref{final_network}, trailer illustrations of the initial and evolved NN topologies and their genetic encoding for the robot-path-planning problem studied here.

% \begin{figure}
% \centering
% \subfigure[Neural Network Topology at the Start of the Optimization Process]{%
% \label{simplearch}%
% \includegraphics[width=0.5\textwidth]{LORD_Set1_of__1ontime54-12_05_Mar_2018.jpg}}%
% \qquad
% \subfigure[Neural Network Topology at the End of the Optimization Process]{%
% \label{complexarch}%
% \includegraphics[width=0.5\textwidth]{lord.jpg}}%
% \caption{Complexification of Neural Network Topology through neuro-evolution --
% \textcolor{red}{**Include the  NEAT string encoding for both images as I had asked you guys to. Also the network topology figures are too small. You have too much white space which should be removed, allowing the figure to use the entire columnwidth**}}
% \end{figure}
Variations of NEAT have seen success in solving complex reinforcement learning problems~\cite{igel2003neuroevolution}, engendering complex behaviors for virtual agents in video games~\cite{stanley2005evolving}, and in developing a crash warning system in automobiles~\cite{kohl2006evolving}. Aside from the artificial intelligence domain, neuro-evolution has also been employed to develop controllers for physical systems such as in multi-legged robot gait control \cite{valsalam2012constructing}. Solutions generated through neuro-evolution have been often shown to surpass those obtained from fixed topology neural networks~\cite{stanley2002efficient}.

\textbf{\textit{Multi-Objective neuro-evolution}}:
Although rare, there exist a few examples of extension of neuro-evolution methods to consider multiple objectives, with applications to autonomous systems~\cite{min2005obstacle} and video games \cite{schrum:aiide08}. Schrum et al.\cite{schrum:aiide08} demonstrated that multi-objective neuro-evolution could be used to generate trade-offs in virtual agents that enable them to quickly and radically change their behaviors to suit the need of the situation. %Furthermore, it was also shown that solutions that required conflicting behaviors could be produced with this approach.

\subsection{Novelty Search \& Experience-Gain}
\label{subsec:novelty}
For autonomous systems, neuro-evolution and its reinforcement learning counterparts\cite{abbeel2007application} are generally associated with training intelligence models where, although the overall operational envelope may be known, the specific operational scenarios are either not known or practically too many to consider. Neuro-evolution deals with this issue by evaluating the performance of each candidate network in the population over a DoE of scenarios in each generation. This strategy often leads to over-fitting, which in this context is referred to as ``deception" \cite{whitley1991fundamental} -- the trained system becomes good at completing tasks for the training (and closely related) scenarios, but often performs poorly when deployed on unseen scenarios (poor at generalization)\cite{risi2009novelty}.

A notable approach to address this issue is \textit{novelty search}, proposed by Lehman and Stanley \cite{lehman2008exploiting}. The fundamental idea was to optimize diversity in the behavioral space instead of performance. Analogical to the evolutionary optimization concept of design-space ``niching" \cite{deb2001multi}, the ``novelty" of each candidate solution was defined as the average distance between that solution and its $K$ closest neighbors in the behavioral space. Within NEAT, this novelty metric was then used as an objective function instead of the corresponding performance.
Lehman and Stanley\cite{lehman2011abandoning} later used this method to solve a maze navigation problem in which three different ANNs were trained through NEAT and compared -- one evolved through novelty search, another evolved through performance search, and a third one evolved via random search. The ANN evolved through novelty search was shown to provide superior performance, by virtue of requiring fewer function evaluations to converge and concurrently achieving a similar performance with a simpler topology.

Since it is practically blind to performance feedback (a core component of adaptation), it is questionable if the benefits of \textit{novelty search} can be generalized or trusted upon. There have been efforts to resolve this question. Cucco and Gomez\cite{cuccu2011novelty} used a weighted aggregate of novelty and performance as the objective function. Another approach \cite{lehman2010revising} defined and treated a threshold performance as a constraint while pursuing novelty search. Mouret and Doncieux \cite{mouret2009using} instead proposed two multi-objective approaches -- called behavioral novelty and behavioral diversity -- and tested their performances on evolving an ANN capable of evaluating a Boolean function with a deceptive fitness. These approaches yielded comparable results to that produced by NEAT in remedying the problem of deception. %\textcolor{red}{**in a single sentence, comment on the performance of their method and the test problem they studied**}.

While borrowing the principle of a pure Pareto search from Mouret and Doncieux \cite{mouret2009using}, the work presented in this paper differs from these aforementioned methods in both how the multi-criteria neuro-evolution is implemented and in the definition of an explorative search criterion called goal-oriented \textit{experience gain}. More specifically, this criterion is designed to serve as a measure of the quantity of \textit{unique local situations} (relative to the environment) that an autonomous system encounters during each mission and is related to its feature application. In other words, this goal-oriented experience-gain does not implicitly contribute to increasing novelty. Instead, it explicitly increases the diversity of training within a threshold time. Therefore this experience-gain increases the learning experience the robot receives, seeking to increase the success in later unseen mission scenarios.
Although the hypothesized outcome of our approach has \textit{conceptual} parallels with exploration in ( $\epsilon$-greedy and softmax-type) reinforcement learning methods  \cite{sutton1998reinforcement,tijsma2016comparing}, exploration is \textit{implicit} as opposed to deliberate (from the perspective of the system) in our case. The \textbf{specific objectives} of this paper can be laid out as:
\begin{enumerate}
	\item To develop an evolutionary approach in designing neural network topologies and connection weights that can simultaneously optimize multiple criteria (e.g., performance on different skills, or performance vs. experience).
    \item To investigate how the performance and training cost of a \textbf{\textit{dual stage neuro-evolution framework}} consisting of a multi-objective neuro-evolution process followed by a single-objective neuro-evolution process) compare with a single-stage single-objective neuro-evolution framework.
    \item To formulate a measure of experience gain for the neural decision-support model, one that is cognizant of mission success and seeks to improve generalization capability.
    \item To investigate the performance of the new neuro-evolution method and its framework variations, by applying them to 2D robot navigation problems, and analyzing the variation in the complexity of the network topology.
\end{enumerate}

The methodology and its application to robot navigation (seeking generalization capability) are collectively named the \textbf{Multi-criteria Evolution of Neural Topologies for Omniscient Robots} or \textbf{MENTOR}. The remainder of the paper is organized as follows: Section \ref{sec:mentor-algo} describes the new multi-criteria neuro-evolution approach; Section \ref{sec:experience-gain} describes the case study, and the new experience-gain and network complexity formulations; Sections \ref{sec:results} presents and discusses the case study results and the evaluation of the evolved ANNs on unseen test scenarios. Finally, concluding remarks are provided in Section \ref{sec:conclusion}.
%
%All of the aforementioned methods combine novelty and performance. These methods use different optimization approaches such as combining the two objectives as a weighted sum, constraining one objective, and using multi-objective optimization. However, these methods implicitly use novelty to capture performance; the use of explicitly defining novelty to obtain performance would serve as a motivation for further research.
%But all these methods try to implicitly capture the desired novelty, while they need a method to define an explicitly desired novelty. CHECK THIS PARAGRAPPH	
%
%The methodology proposed in the following research changes this idea by defining a goal-oriented experience. In this approach, performance is optimized against useful experience gained, referred to as \textit{experience-gain}, during the training. While novelty itself might be irrelevant to the desired performance, experience-gain, as defined in this research, is defined based on the desired performance. By adding an implicit constraint, the experience is defined in a way that it guarantees that the experience-gain is relevant to the desired performance. Still, the proposed formulation allows the candidate solutions to search in space that may not lead to the best performance and avoids premature convergence.
\vspace{-0.5cm}
%%%%%%%%%%%%%%%%%%%%%%%%%%%%%%%%%%%%%%%%%%%
%================MONEAT===================
%%%%%%%%%%%%%%%%%%%%%%%%%%%%%%%%%%%%%%%%%%%
\begin{figure*}
\vspace{-1.5 cm}
\centering
\includegraphics[page=3,width=1.4\columnwidth]{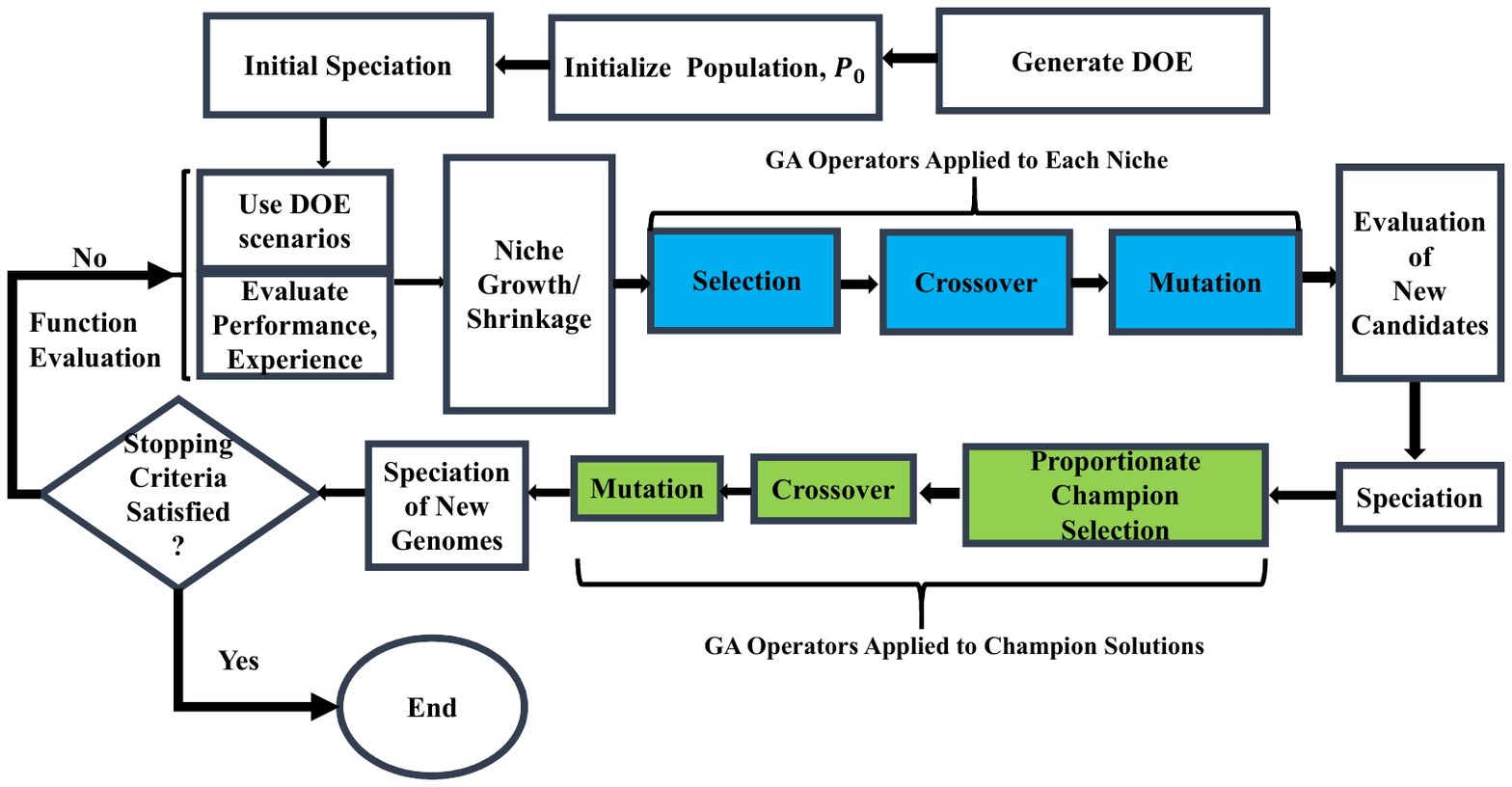}
\captionsetup{justification=centering}
\vspace{-1.5cm}
\caption{MENTOR: Multi-Criteria Neuro-Evolution Algorithm}
\label{fig:flow_all}
\end{figure*}

%%%%%%%%%%%%%%%%%%%%%%%%%%%%%%%%%%%%%%%%%%%%%%%
%==================MENTOR======================
%%%%%%%%%%%%%%%%%%%%%%%%%%%%%%%%%%%%%%%%%%%%%%%
\section{MENTOR: NEURO-EVOLUTION ALGORITHM}\label{sec:mentor-algo}
\subsection{Genetic Encoding of Neural Networks}
Here we extend the original NEAT (neuro-evolution of augmenting topologies) algorithm \cite{stanley2002evolving,stanley2002efficient,stanley2005evolving} to a multi-criteria search process through an elitist non-dominated sorting selection strategy \cite{deb2002fast} and a modified niching/speciation approach. A flowchart of our new multi-criteria NEAT algorithm is shown in Fig. \ref{fig:flow_all}. % (the colored blocks elucidate steps where we particularly differ with elucidating the key distinctions between our implementation and the original implementations of NEAT.
Description of the key components of the algorithm, particularly focusing on those where we differ from the original NEAT, is provided below.

In a similar vein to the direct encoding used in NEAT, genes encode the edges of a directed graph, and genetic operators alter these genes. Each gene consists of an origin node, a terminal node, the associated weight of the edge, and a quantity termed as the innovation number, as depicted in Fig. \ref{final_network}. The innovation number of a gene indicates its heredity and chronological order of creation (over generations), which is used during the crossover.
% . When comparing or recombining two ANN candidates, if one gene is present in both networks' genotype, these genes are referred to as matching genes. If a particular gene does not match between the two networks' genotypes, it is termed as either a disjoint gene or an excess gene. Excess genes are genes that are created in one neural network after the creation of the latest gene in the other neural network.

\subsection{Speciation and Selection}
%In this paper, a multi-objective solution through the use of neuro-evolution has been implemented by combining rank-based non-dominated sorting which is used in other optimization methods such as the non-dominated sorting genetic algorithm (NSGA - II) developed by Deb et al.\cite{deb2002fast} with the existing implementation of NEAT, along with making a few changes to both of the disparate algorithms.
%In this approach comparison is done based on the rank the candidate solutions.
Similar to NEAT \cite{stanley2002evolving}, we perform hierarchical reproduction -- i.e., in each generation, selection, crossover, and mutation are first performed within each species, and then champions from each species are combined and put through these reproductive operations. In NEAT, both the topology and the weights of the neural network change; changes in the topology usually need time (multiple generations) to train their weights. Therefore complex networks might be discarded by selection pressure before they find appropriate weights. Speciation, or preservation of species, assists newly created neural topologies in surviving competition with simpler/older topologies with relatively more stabilized (fewer) connection weights. Speciation can be likened to clustering of the different genomes. The population of candidate ANNs is divided into species or groups, by comparing each ANN with others in terms of the following distance metric:
\begin{equation}
D_{i,j}=\frac{c_1 E}{N}+\frac{c_1 J}{N}+c_3 \bar{W}
\label{distmet}
\end{equation}
In Eq. \ref{distmet}, $D_{i,j}$ is the distance between two candidate ANNs, $i$ and $j$; $E$ is the number of excess genes and $J$ is the number of disjoint genes; $\bar{W}$ is the summation of the weights difference between matching genes; $[c_1,c_2,c_3]=[1,1,0.4]$ are user defined coefficients. Each child genome, after its creation, is associated with a particular species, based on the distance metric. If none of the niches are suitable (i.e., roughly speaking, genetically unrelated), the new genome is allowed to create its own niche.

The fitness evaluation and selection process differ from original NEAT to enable the \textit{multi-criteria} search capability. We adopt the principles of elitist non-dominated sorting \cite{deb2002fast} to assign ranks and create the mating pool. We use tournament selection in intra-species reproduction, and rank-based proportionate selection is used in global reproduction.

%\subsection{Speciation}
 %This will help these new genomes to have time to grow and train. Also, niching helps exploit one region better. In the algorithm used in this paper, genomes in one niche are selected to perform GA operations inside the niche.

\subsection{Niche Growth and Shrinkage}
In this step, the size of (i.e., allowed number of members in) each species or niche is regulated in order to preserve diversity. The updated size is computed by first adjusting the fitness with niche count, computing the average fitness of each niche, and scaling it with the average fitness across the population. The species size formulation can thus be expressed as
\begin{equation}
{N_{i}'}=\frac{\sum_{j=1}^{N_{i}} \bar{F_{i,j}}}{\bar{F}}
\label{growshrinknorm}
\end{equation}
%
%\textcolor{red}{it's not clear from this equation how niching leads to the following growth/shrinkage phenomena, described below.}
% In this approach, the first step is to adjust the fitness by niching count and then find the average adjusted fitness of one niche compared to the whole population.
% Eqs. \ref{adjustbynichcount} , \ref{deltanichingcount} illustrate this procedure.
% In this equation, $d_{i,j}$ is the distance between two points in the design space. Eq. 1 is used to estimate the distance between two neural networks.

% \begin{equation}
% {F_{i}}=\frac{f_i} {\sum_{j=1}^{N} \delta_{i,j}}
% \label{adjustbynichcount}
% \end{equation}

% \begin{equation}
% \delta_{i,j}=\{\begin{tabular}{*{2}{c}}
% $0$ & if $d_{i,j}>\Delta$ \\

% $1- (\frac{d}{\Delta})^{\alpha}$ & if $d_{i,j} \leq \Delta$
% \end{tabular}\}
% \label{deltanichingcount}
% \end{equation}

In Eq. \ref{growshrinknorm}, $N_{i}$ and $N{i}'$ are the original and updated sizes of the $i^{th}$ niche $i$ respectively. Here, $ \bar{F_{i,j}}$ is the adjusted  fitness of the $j^{th}$ candidate in the $i^{th}$ niche (where the original fitness derived from rank is adjusted using the crowding distance approach \cite{deb2001multi}), normalized by the size of that niche; $\bar{F}$ is the average adjusted fitness of the whole population normalized by the population size.
If the current size of a niche is smaller than its desired size (by virtue of its fitness), some candidate genomes in the niche will undergo mutation to create a new population. If the current size is greater than the desired size, some genomes are discarded.

% Based on this formulation, the size of each niche is controlled by overall average, and therefore its growth is limited. This will lead to maintaining more diversity in the population.

%Therefore even if one species performs very well, niching will lead to its shrinkage in order to stop the entire population from being generated from that niche. On the other hand, a niche which does not perform well will have a greater chance to remain as long as they are diverse. Therefore this procedure will discard some genomes, even though they have good performance, in order to retain a more diversified population. %For a maximization problem, the inverse of rank is used instead of the objective function. The rank is used instead of an objective function for a minimization problem.
% \vspace{-0.5cm}
\subsection{Crossover and Mutation}
The crossover approach used here is adopted from the original implementation of NEAT: two parents genomes are compared, and the matching genes from each parent are added to the children randomly; the excess and disjoint genes are only added to the children from the parent with the better rank (or randomly, if both parents belong to the same rank).

% The only difference is that instead of objective function the rank is compared between two parents.
%In multi-objective representation, the rank is used instead of objective function.
%\subsection{Mutation}
Three different mutation operations are applied to each candidate ANN. The first two operates on each edge, while the third mutation operation is applied to different pairs of nodes. More specifically, the first operation changes the weight of an edge, the second operation adds a node in the middle of an existing edge and divides it into two edges, and the third operation adds an edge between two existing nodes. After adding an edge or a node, which leads to the creation of two new edges, each new edge receives its innovation number. We apply probabilistic mutation on each edge (fixed probability of mutating each edge of a candidate network, albeit a smaller probability value). This is done to make the likelihood of mutating the network proportional to the network's topological complexity (e.g., adding one node to a 100-node network entails a different degree of variation compared to adding a node to a 10-node network).

% \subsection{The neuro-evolution Procedure}
% In the procedure used for this endeavor, a GA is first applied to each niche to improve its performance. After applying the GA on all the niches, the best candidates from each niche are used in another GA.
% Figure(\ref{fig:flow_all}) illustrates this procedure.
% \begin{figure}[H]
% \centering
%   \includegraphics[scale=0.3, height=8cm, width = 0.5\textwidth]{flowProcedure.JPG}
%   \caption{Flowchart Depicting the Optimization Procedure}
%   \label{fig:flow_all}
% \end{figure}
% Table\ref{tab:one} summarizes the most salient parameters that were used the for multi-objective NEAT optimization in this paper.

%\textcolor{red}{it would be a good idea to include the commented out table, to summarize the user-prescribed parameter values that you used}
%\vspace{-0.5cm}
%================Simulation===================
\section{MENTOR: OPTIMIZATION FORMULATION}\label{sec:experience-gain}
%\vspace{-0.5cm}
\subsection{Application \& Design of Experiments}
\label{subsec: DOE}
\begin{figure}[h]
\centering
\includegraphics[width=0.4\textwidth,angle=270,origin=c]{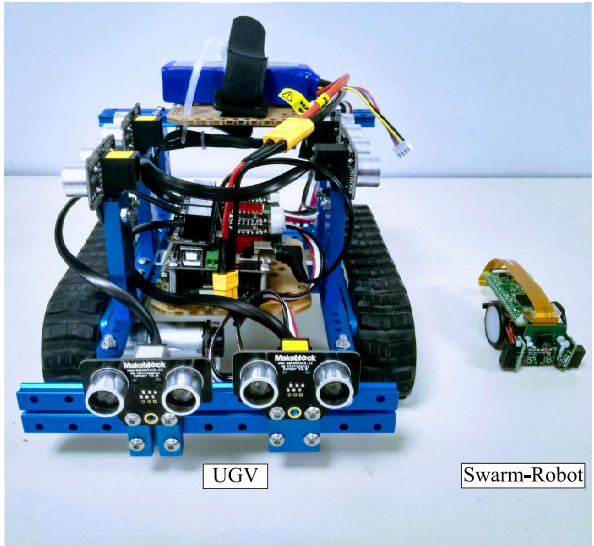}
\captionsetup{justification=centering}
\caption{Robotic Platforms that are Tested in Simulation}
\label{robots}
% \vspace{-0.5cm}
\end{figure}

%\vspace{-0.5cm}
Although the ``experience-gain" concept (introduced in Section \ref{subsec:novelty}) can be used in different application contexts, it will be helpful to first provide the backdrop of the typical 2D robot navigation problems studied in this paper. We are designing a neural network-based path planning system that allows a small ground robot to autonomously navigate from a source to a target location while avoiding stationary obstacles (all within a simulated environment). We perform our algorithm evaluation and analysis on two different robots: 1) A small indoor UGV, (to be referred to as \textbf{UGV} here onwards) equipped with eight proximity sensors -- placed in pairs ($0.075m$ apart) on each of its four edges; and 2) an \textbf{original robot designed in our laboratory for swarm applications\cite{swarm_bot}} (to be referred to as \textbf{Swarm-Robot} here onwards), of size $8cm\times 6cm$, and equipped with three proximity sensors -- all of which are placed on the front edge of the robot as shown in Fig \ref{robots}. By testing our methodology on these two robots in a virtual environment, we are also able to observe and discuss the effects of input space dimensionality (the readings from the proximity sensors serve as inputs to the ANN -- refer Section \ref{subsec: SA}) on the duration of the training procedure.

%The proposed intelligent system was trained with two separate procedures. In the first procedure, two-stage training was conducted with the first stage comprising a multi-objective optimization stage and the second stage is a single objective optimization stage. For the second procedure, training was only performed via single objective optimization.
% \begin{figure*}
% \vspace{-0.5cm}
% \centering
% \begin{tabular}{c c}
% \includegraphics[width =0.3\textwidth]{robot1.pdf} &
% \includegraphics[width =0.3\textwidth]{robot2.pdf}\\

\begin{figure}[h]
\centering
\includegraphics[width=0.5\textwidth]{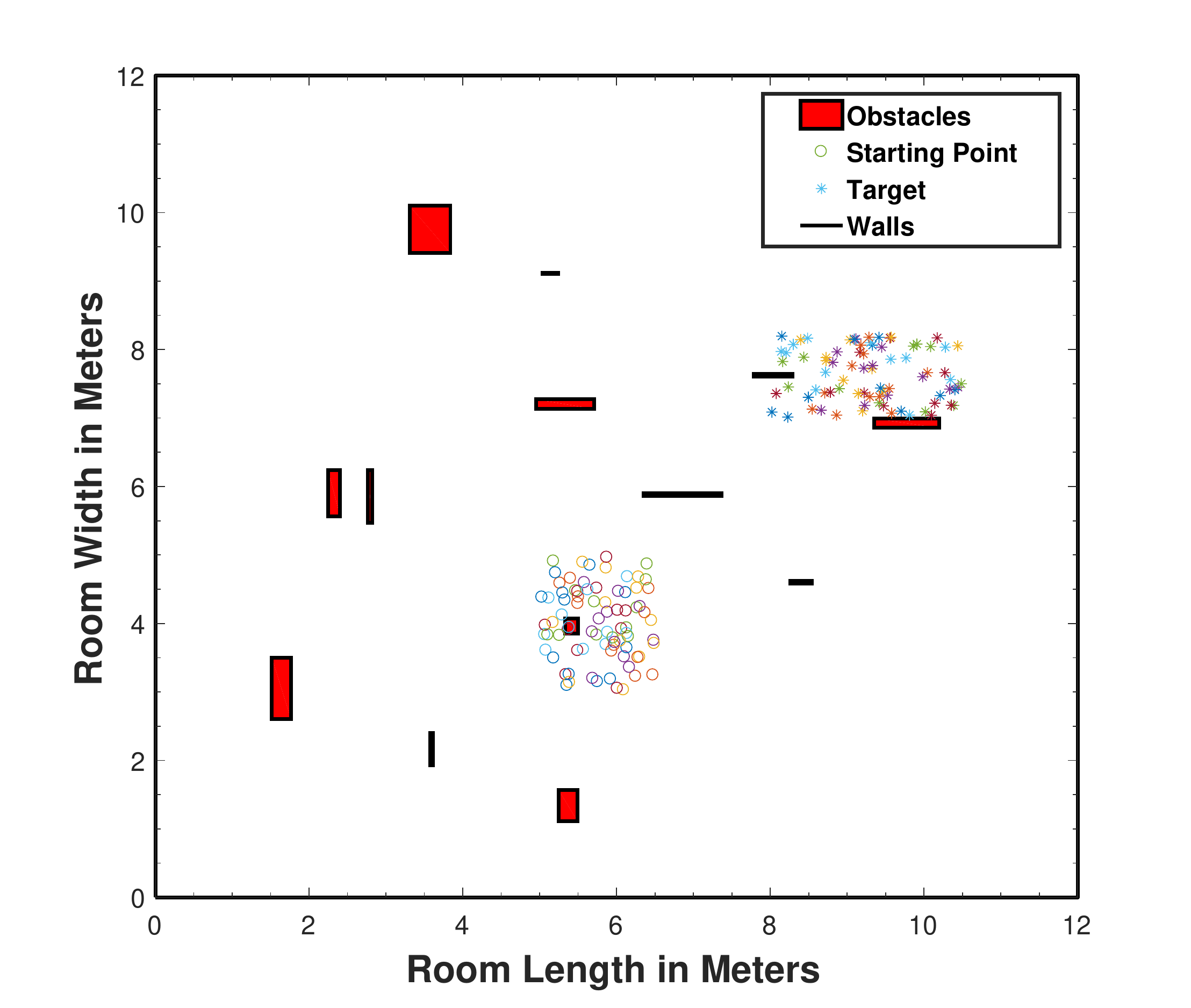}
\captionsetup{justification=centering}
\caption{Virtual Environment for Evaluating the ANN-based Path Planner (showing obstacles and start/target locations)}
\label{env2}
% \vspace{-0.5 cm}
\end{figure}

% (a) UGV & (b) In-house Designed Swarm-Robot
% \end{tabular}
% \captionsetup{justification=centering}
% \caption{Robotic Testbed Platforms}
% \label{robots}
% \vspace{-0.5 cm}
% \end{figure*}
During neuro-evolution, each candidate neural network system is evaluated on a set of sample scenarios generated via design of experiments. %The primary-stage multi-objective optimization training was performed on a composite environment with $20$ randomly chosen to start and to end locations; the second-stage single-objective training was conducted on the same composite environment with $200$ randomly chosen to start and to end locations.
The mission scenarios are generated as source-target pairs in a composite environment comprising areas with crowded and sparsely distributed obstacles. The floor area of the square environments is set to $12m \times 12m$ for the UGV and $5m \times 5m$  for the Swarm-Robot. To generate the obstacles, the environment is divided into $4$ grids. Latin hypercube sampling is performed on each grid to generate the coordinates of the centroid of each obstacle and size of the rectangular obstacles. A constrained sampling approach is taken to ensure non-overlapping obstacles. Figure \ref{env2} illustrates one such environment that was employed to evaluate the ANN-based path planner.
\subsection{State and Action Spaces}
\label{subsec: SA}
\begin{figure}[h]
\vspace{-1 cm}
\centering
\includegraphics[page=2,width=0.5\textwidth]{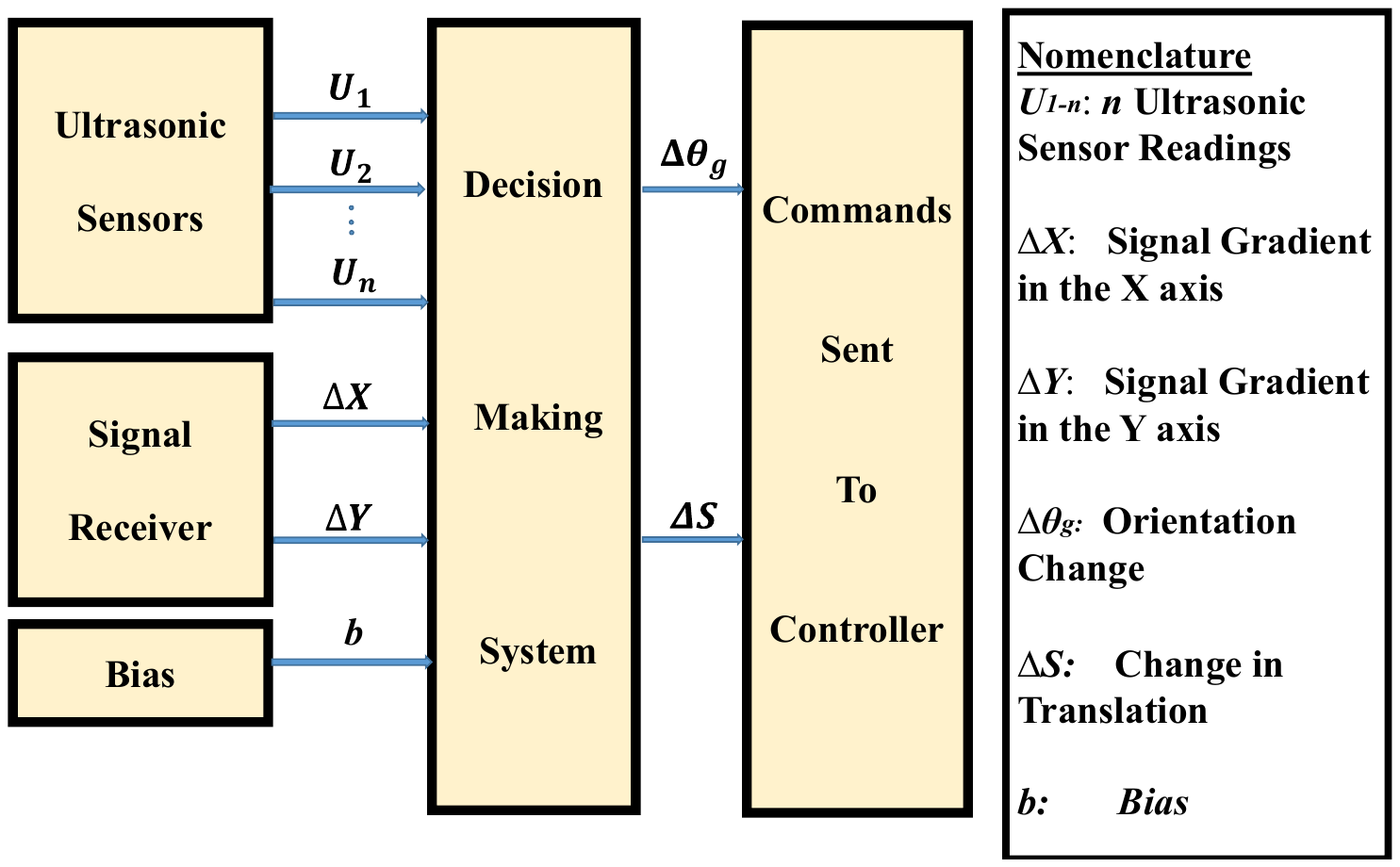}
\vspace{-1cm}
\captionsetup{justification=centering}
\caption{State-Action Spaces of the ANN-based Path Planning System}
\label{fig:ActionStates}
% \vspace{-1 cm}
\end{figure}

The state and action spaces of the robot directly represent the inputs and outputs of the neural network-based path planner that decides the next local action to take given the current state of the robot in the environment. Figure \ref{fig:ActionStates} summarizes the inputs and outputs of the neural network. The state space of the robot include the sensory readings, namely the 8 (or 3 for the Swarm-Robot) distances, in meters, produced by the ultrasonic sensors; a signal-strength term, which is defined as the reciprocal of the distance to the goal, from which, two signal gradients are defined as the rate of change of this signal with respect to change in the distance to the goal during one time step; the current global pose of the robot constitutes another input while the bias of the network constitutes the final input. Thus we require ANNs with 11 inputs for the UGV and ANNs with 6 inputs for the Swarm-Robot. The action space corresponds to the movements the robot is capable of making: the first representing the angle, in radians, through which the robot must rotate at its current point and the second representing the distance, in meters, through which the robot must translate after rotation. The first output is restricted to take values between $-\pi$ to $\pi$ for both robots while the second output is constrained to take values between $0.1 m$ to $2m$  for the UGV and $0.01cm$ to $0.08cm$ for the Swarm-Robot; both outputs are continuous variables. The output of the neural network considers only the next immediate action, thus making it akin to the myopic reinforcement learning paradigm.

 %Here $U_{1}$ to $U_{8}$ represent the eight proximity sensor readings, $\Delta X$ and $\Delta Y$ denote the signal gradients in the X and Y directions respectively, while $\theta_{g}$ refers to the global heading of the system. Similarly, $\Delta \theta_{g}$ and $\Delta S$ refer to the change in the global heading and the translation to be made at the next time step respectively.
 %

\subsection{Criteria Functions}
Our objective in posing the aforementioned path planning problem as a multi-objective search process is to allow balancing (exploitative) performance with (explorative) experience. Typical of the problem used here, performance can be defined in terms of the time taken or the closeness to the target location achieved by the end of a fixed simulation time. \textit{Experience-gain} can be perceived as the number of different types of local environment encountered and responses of the robot to those environment. Maintaining ``survivability" (which, in this context explicitly refers to the ability of the robot to avoid colliding into obstacles) and allowing mission-completion-success to be achieved are important considerations here, that differentiates our work from earlier novelty search implementations\cite{lehman2011abandoning}. Our approach is designed to allow useful novelty to be incorporated in the neuro-evolution process. By pursuing a multi-objective neuro-evolution, loosely speaking, some candidate ANNs will aggressively evolve to achieve better performance function values (reaching the target as fast as possible), while some candidate ANNs will implicitly deviate from potential shortest paths to explore diverse obstacle-encounter scenarios, automatically allowing them to test their survivability (without compromising mission success) w.r.t. diverse local environments. Mixing of the genetic schemas of these ANNs is hypothesized to lead to ANNs that embody the best of both qualities (with different trade-offs), and thus more capable of navigating environments beyond what they have experienced during the evolution-based training process. Assuming a \textbf{max-max} problem, the two objective functions are specifically called \textbf{Performance} and \textbf{Experience-Gain}. Before delving into the descriptions of these criteria functions, lets delineate the terminology used in defining these functions.%Accordingly, a two-stage optimization problem was formulated. In the first stage, a \textit{max-max} multi-objective optimization problem was formulated with the following objective functions:

% \begin{equation*}
% \begin{aligned}
% & \underset{x}{\text{maximize}}
% & & F(x) \\
% & \text{subject to}
% & & f_i(x) \leq b_i, \; i = 1, \ldots, m.
% \end{aligned}
%\end{equation*}
\begin{itemize}
\item $T_{tot,i}$: The total time it would take the robot to reach the target or goal, in the $i^{th}$ scenario, if it were to take the hypothetical straight line path from its starting position (disregarding obstacles), with the robot moving at a rated average velocity.
\item $T_{rem,i}$: The remaining time when a simulation ends, in the $i^{th}$ environment. By providing this parameter, it is ensured that robots are provided with an incentive to head towards the goal rather than cycle around aimlessly.

\item $\delta_{i}$: the success or failure in one specific scenario.
\item $S_{f,i}$: the distance recorded by the robot at the final
instant of time in a simulation run in the $i^{th}$ scenario.
\item $F_{i}$: the objective function value of an ANN resulting from robot simulation in the $i^{th}$ scenario.
\item $G_{i}$: the total number of relatively unique points experienced by the robot during its simulation in the $i^{th}$ scenario. %\textcolor{red}{How do you mathematically define unique points? This is very important to describe with equations.}
\item $D_{final,i}$: the displacement at the final time step with respect to the initial pose during the $i^{th}$ scenario.
\item $\alpha$: A user defined \textit{importance} factor which indicates the relative importance of performance w.r.t gaining experience. This can also be being perceived as the relative importance given to exploitation while the system performs exploration. This heuristic factor was set at $0.2$.
\item $n$: Total number of training scenarios.
\end{itemize}

\textbf{Performance}: The first maximization criteria function drives the robot to head towards a particular target, i.e., the performance indicator.
% A signal strength term was defined to help quantify our performance metric.
This objective function is denoted by $F$, and defined in Eqs. \ref{sum:fitness1} and \ref{indfitness1}. The first term of Eq. \ref{indfitness1} indicates whether or not the robot managed to reach its destination within the pre-defined simulation time; the second term is a function that penalizes the final distance of the robot from its destination.
%Here $w_{1}$ and $w_{2}$ are the weights of the time and distance terms in objective function.
% \vspace{-0.5cm}
%
\begin{equation}
F = \sum_{1}^{n} F_{i}
\label{sum:fitness1}
\end{equation}
%
% \vspace{-0.5cm}
\begin{equation}
F_{i} = \delta_{i}\times(1+\frac{T_{rem,i}}{T_{tot,i}} ) + (1-\delta_{i}) \times \frac{1}{1+S_{f,i}}
\label{indfitness1}
\end{equation}
%
% \vspace{-0.5cm}
% \begin{equation}
% F_{i} = w_{1}\times\frac{min(S_{max,i},S_{end,i})}{S_{max,i}}+w_{2}\times\frac{T_{rem,i}}{T_{tot,i}}
% \label{indfitness1}
% \end{equation}
% %
% \vspace{-0.5cm}

Here $T_{rem},T_{tot}$  are the remaining time when the robot reaches the goal and the total time of the experiment respectively. $S_{f}$ is the distance to the target in the last step.
$\delta_{i}$ is $1$ if robot reaches the goal and it is $0$ if it cannot reach the goal.

Based on the value of $\delta_{i}$, if robot reaches its goal the objective function, given by Eq. \ref{indfitness1}, is dictated by how quickly the robot reached the destination; and if it fails, the same objective function is dictated by the final proximity of the robot to the goal.
Both $\frac{T_{rem,i}}{T_{tot,i}}$ and $\frac{1}{1+S_{f,i}}$ are between $0$ and $1$. Therefore $1+\frac{T_{rem,i}}{T_{tot,i}}$ is always greater than $\frac{1}{1+S_{f,i}}$, which means that if the robot reaches the goal its performance value will always be better than the robot which did not reach its destination.

\textbf{Experience-gain}: This second maximization criteria function incentivizes the robot to capture more information regarding its surroundings. An \textit{experience point}, in this context, has three components: the state of the robot before it executes an action, the action it takes, and the robot's state after applying said action. Different scenarios cause different experiences, and the aggregate experience of the robot must include these experiences. This approach is encapsulated in Eq. \ref{uninon_exp}.

\begin{equation}
E=\bigcup _{i=1}^{N_s}  E_{i}
\label{uninon_exp}
\end{equation}
Eq. \ref{uninon_exp} dictates that the overall experience-gain is the union of all the different experiences encountered by the robot. However, it has to be noted that although each experience point is ``unique'', collectively speaking, the set lacks diversity as many of these experience points might represent repeated experiences.

%Eq. \ref{uninon_exp} dictates that the union of experience points is used in order to exclude the effect of repeated experiences. Although all of the experiences in $G$ are unique, they do not have diversity. It is more reasonable to decrease the effect of similar experiences.

The task of removing similar experience is not trivial. Although some experiences have similarities, they should not be completely removed. A suitable analogy for this problem would be a complete bidirectional graph: each experience point can be considered to be a node, and each node is connected to one another through an edge -- the edge length between two nodes represents the distance of the corresponding experience points. Thus, each experience unit can be represented as:

\begin{equation}
W_{e_{i,j}}= |V_i-V_j|
\label{weight_of_graph}
\end{equation}
where $V_i, V_j$ are vectors in experience space, which can be expressed as:
% \vspace{-0.5cm}
\begin{equation}
\begin{aligned}
V_{i}= {} & [U_1(t), ... , U_{N_{US}}(t), A_1(t),...,A_{N_{A}}\\
& (T),U_1(t+1),..., U_{N_{US}}(t+1) ]^T
\label{Vertex_exp}
\end{aligned}
\end{equation}
Therefore each node is actually a point in a ($2 \times N_{US} +N_{A}$) dimensional space. This ($2 \times N_{US} +N_{A}$) dimensional space is called the Experience Space in this paper. Here,  $N_{US}$ and $N_{A}$ are the number of ultrasonic sensors and actions, respectively.

The overarching goal of this second criteria function is to, quantitatively speaking, define the unique experiences collected by the robot during its missions. To this end, the aforementioned graph analogy is used to develop and provide a better understanding of the concept of \textit{experience}. More concretely, the notion of minimum spanning trees (MST) was adopted in this paper to simultaneously capture the uniqueness and overall coverage of the experience gained by the robot.

% An MST is a subgraph of a connected weighted graph which possesses the following features:
% \begin{enumerate}
% \item Possesses no cycles -- an MST is a tree and hence has no cycles or circuits.
% \item Contains or \textit{spans} all the nodes of the original graph
% \item The sum of edge weights is minimum -- i.e., among all spanning trees constructed from the original graph, the MST possesses the minimum aggregate edge weight.
% \end{enumerate}
There are different approaches to find an MST -- Kruskal's algorithm \cite{kruskal1956shortest} is used in this paper. The complexity of the algorithm is $O(E \log(E))$, where $E$ is the total number of edges. A description of the algorithm can be found in Algorithm \ref{Kruskal} in the Appendix. The overall experience-gain criterion function, $G$, can thus be expressed as

% \vspace{-0.5cm}

\begin{equation}
G = \sum_{e_{i} \in MST} W_{e_{i,j}}
\label{indfitness2}
\end{equation}
It can be seen from Eqs. (\ref{sum:fitness1}) and (\ref{uninon_exp}), that the objective functions to be used during neuro-evolution is given by considering the performance and experience-gain criteria functions over all the scenarios in the DoE.

%Finally, terminologies used in Eq.(\ref{indfitness1}) and Eq.(\ref{indfitness2}), refer to:
%Eq.(\ref{indfitness1}) and eq.(\ref{indfitness2}) represent the two objective functions for a single environment. The final objective functions are obtained by summing the two objective functions separately over all the mission scenarios.

%\textbf{\textit{Experience-Gain Formulation}}: Experience-gain, for the 2D navigation application used in this paper, encapsulates the obstacle information collected by the system via its proximity sensors. %\textcolor{gray}
%Note: In this draft manuscript, owing to time constraint, we used a preliminary experience-gain formulation; scope for improvements remains, with regards to using better measures of sample uniqueness to define unique experience-gain -- to be included in the final manuscript. Experience-gain here is defined as (aggregated) significant difference in the ultrasonic sensor readings between consecutive time-steps, with a threshold $\epsilon$ (set through empirical tests) determining significance. The aforementioned experience-gain formulation is algorithmically explained in algorithm \ref{algorithm:two} in the Appendix.

\subsection{Case Studies: Two-Stage vs. Single-Stage}
\label{subsec:Case_studies}
%Two different case studies are undertaken in this paper: both these case studies involve training and testing two neural networks on the two robots described in Section \ref{subsec: DOE}.
Two different frameworks are implemented to explore the performance of single and multi-objective neuro-evolution.
\begin{enumerate}
\item \textit{Dual stage optimization process}: A multi-objective optimization, maximizing both \textbf{performance} and \textbf{experience gain}, with a pre-defined number of generations is first performed, constituting stage 1. A portion of the final population of the multi-objective phase is used as the initial population for a single-objective neuro-evolution process that only maximizes \textbf{performance}, constituting stage 2. In stage 2, all other aspects of the MENTOR algorithm are persevered except for the selection process. Instead of non-dominated sorting, a direct rank based proportionate selection is used.
\item The second network is trained via a single stage, single-objective neuro-evolution process.
\end{enumerate}
For the sake of brevity, application of the above frameworks to the UGV and the Swarm-robot will be respectively denoted as \textbf{Case - UGV} and \textbf{Case - Swarm-Robot}. To ensure fair and meaningful quantitative comparison of the two optimization frameworks, an equal number of training scenarios are used for both optimization framework, and an equal number of maximum function evaluations are allowed for both frameworks.

The emphasis of this study is on how topological diversity and complexity varies under the two approaches, and on whether experience-gain promotes performance beyond the evolutionary training period. Furthermore, by performing these studies on two different robots, involving different numbers of sensors, we are able to shed light on the effect of the dimensionality of the input space on the rate of evolutionary learning.

\subsection{Computational Complexity of ANN Topologies}
An important aspect of neuro-evolution algorithms is their ability to evolve different topologies of neural networks and preserve topological complexification through speciation and niching. Although this unique characteristic of neuro-evolution is well known, there is a lack of methodical/quantitative analysis of the computational complexity of the candidate ANNs produced and preserved by the evolutionary process. %To address this gap, in this paper, we derive a measure of ANN complexity, with the premise that topology with greater complexity is capable of capturing more information.
To address this gap, in this paper we develop a preliminary complexity measure that estimates the approximate run-time complexity of candidate neural network topologies. More specifically, we consider the approximate number of floating point computations or FLOPS required by the candidate network topology to convert the input vector into the output (only an approximation is possible due to the subjective nature of the complexity of special (nonlinear activation) functions, e.g., $tanh$, within ANNs). Both the architectural complexity of the network (number of neurons and inter-neuron connectivities) and the nonlinearity of the activation function are considered, and the estimated run-time \#FLOPs of the given candidate network is scaled by that of the simplest possible network topology (where the simplest network only includes input and output neurons, and no hidden neurons).

\section{MENTOR: ANALYSES AND DISCUSSION}\label{sec:results}
\subsection{Implementation}
To program and implement MENTOR in the virtual environment (including modeling the robot kinematics and sensor behavior), MATLAB\textsuperscript{\tiny\textregistered} is used. The parameter settings that are used in the Case Studies are summarized in Table \ref{parameters for GA}. Research data and codes related to MENTOR implementation can be found at {\scriptsize\url{http://adams.eng.buffalo.edu/algorithms/neuroevolution/}}.

\begin{table}[h]
% \vspace{-0.5cm}
\caption{Neuro-evolution Parameters}
\label{parameters for GA}
\begin{tabular}{*{3}{c}}
\toprule
Parameter & Case - UGV & Case - Swarm-Robot\\
\midrule
Number of Genomes & 100 & 100\\
Maximum Iteration & 50 & 50\\
Elitist Preservation  & $2\%$ & $2\%$\\
Crossover Probability & 0.85 & 0.85\\
Weight Mutation  & $P=0.25$ & $P=0.25$\\
Node Addition Mutation & $P=0.05$ & $P=0.08$\\
Edge Addition Mutation & $P=0.03$ & $P=0.5$\\
\bottomrule
\end{tabular}
\end{table}

%==================Subsection========================================
\subsection{Results: Case - UGV}
\label{sec:Results}
As mentioned in Section \ref{subsec:Case_studies}, there are two optimizations carried out to obtain two distinct neural networks. For both approaches, $80$ training scenarios are used.

\begin{figure}[h]
%\vspace{1 cm}
\centering
\includegraphics[width=0.5\textwidth]{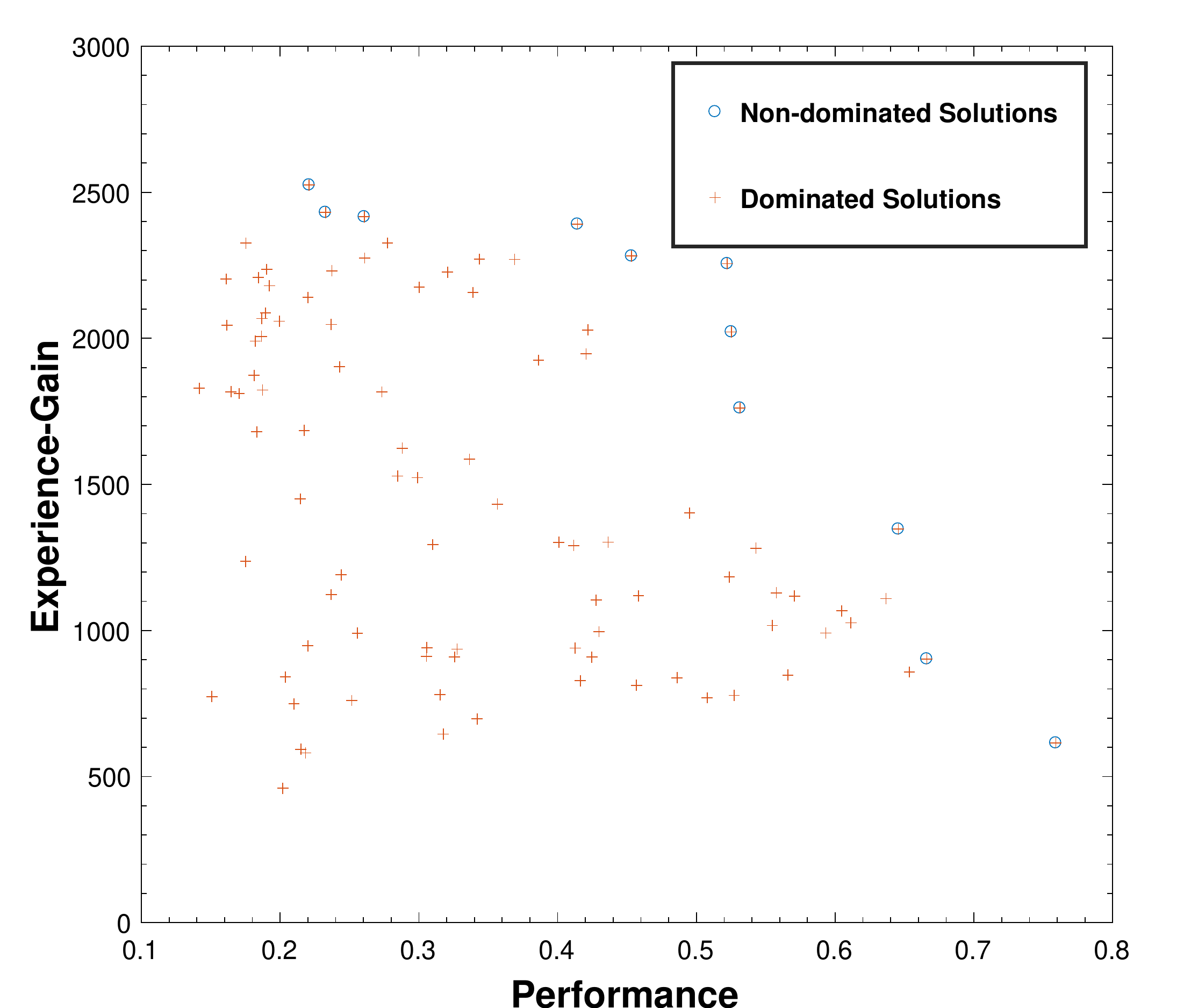}
\captionsetup{justification=centering}
\caption{Case - UGV: Pareto Front of the Multi-Objective Optimization Process for Dual stage}
\label{ParFr:Large}
% \vspace{-0.5cm}
\end{figure}
%\vspace{-1cm}
Figure \ref{ParFr:Large} shows the Pareto front obtained by the multi-objective optimization of the dual stage framework. The significant trade-offs between performance and experience gain is readily evident from Fig. \ref{ParFr:Large}.

\begin{figure}[h]
\centering
\includegraphics[width=0.5\textwidth]{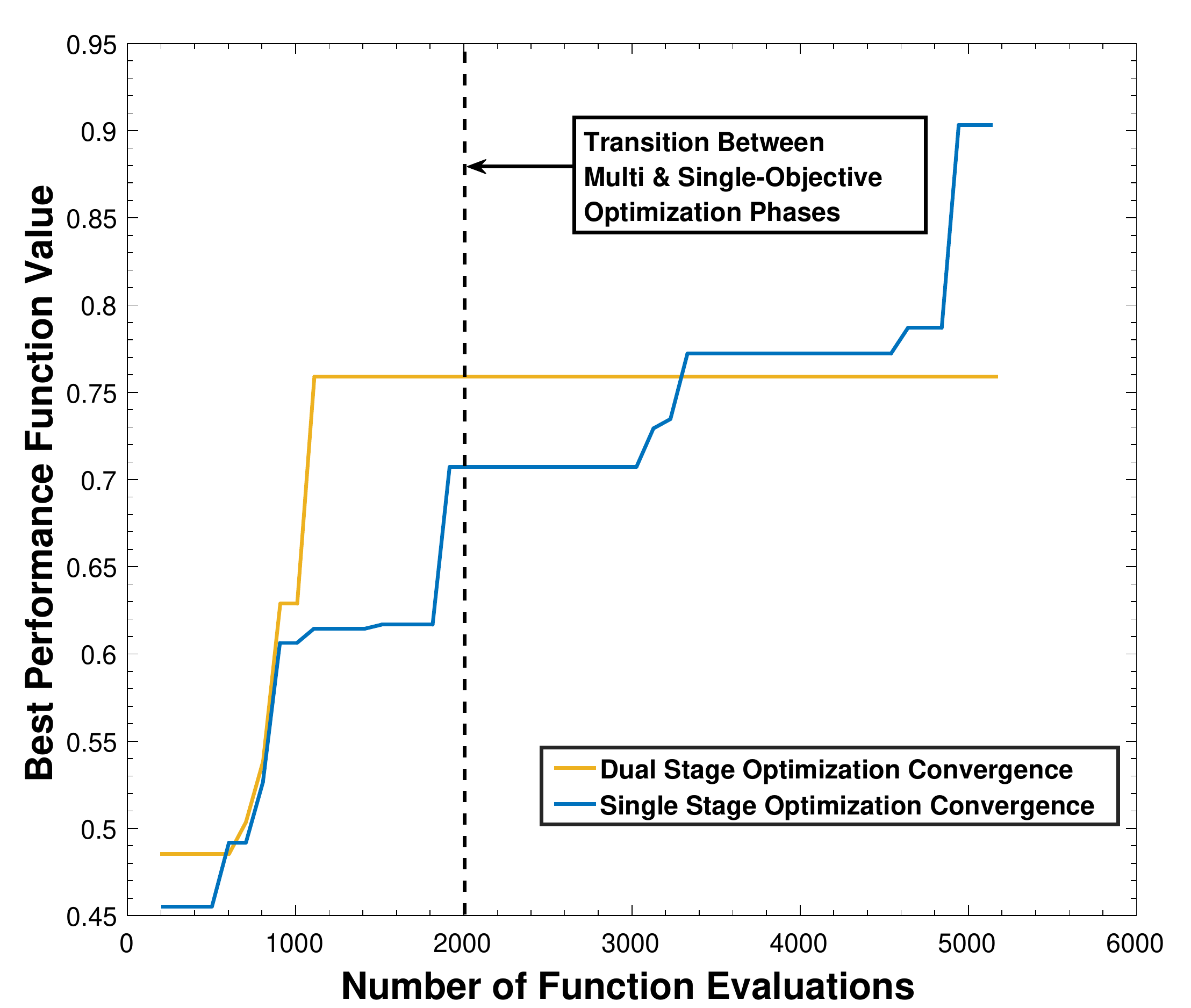}
\captionsetup{justification=centering}
\caption{Case - UGV: Comparison of Convergence Histories}
\label{conv_hist:Large}
\end{figure}

Figure \ref{conv_hist:Large} shows the convergence history of both dual-stage and single-stage optimizations for the UGV. Although the single-stage optimization outperforms the dual-stage optimization process eventually, it is interesting to note that the dual-stage optimization provides a faster rate of increase in performance.

\begin{figure}[h]
\centering
\includegraphics[width=0.5\textwidth]{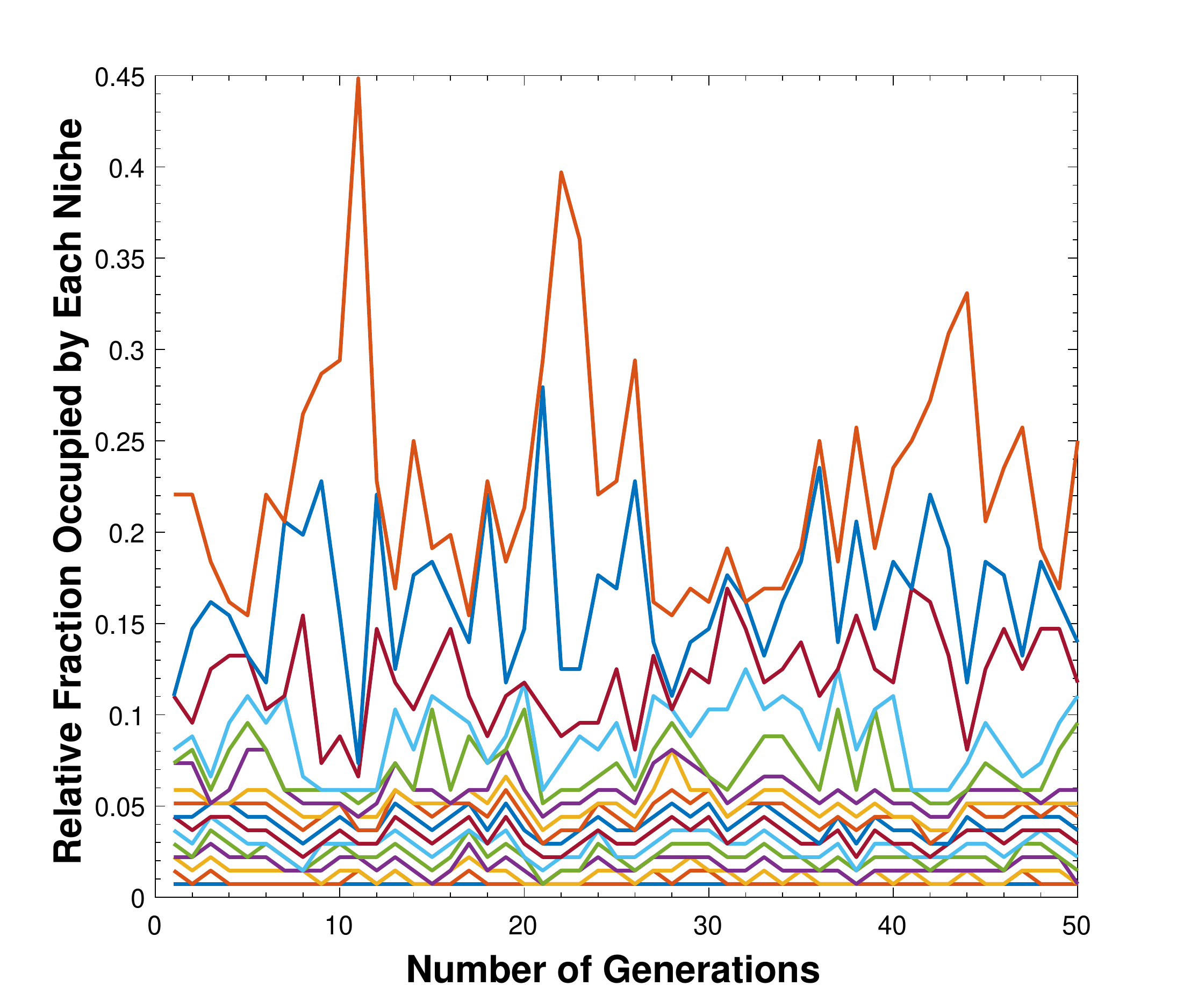}
\captionsetup{justification=centering}
\caption{Case UGV: Dual Stage Variation of Niches}
\label{var_multi:Large}
% \vspace{-0.5cm}
\end{figure}
\begin{figure}[h]
\centering
\includegraphics[width=0.5\textwidth]{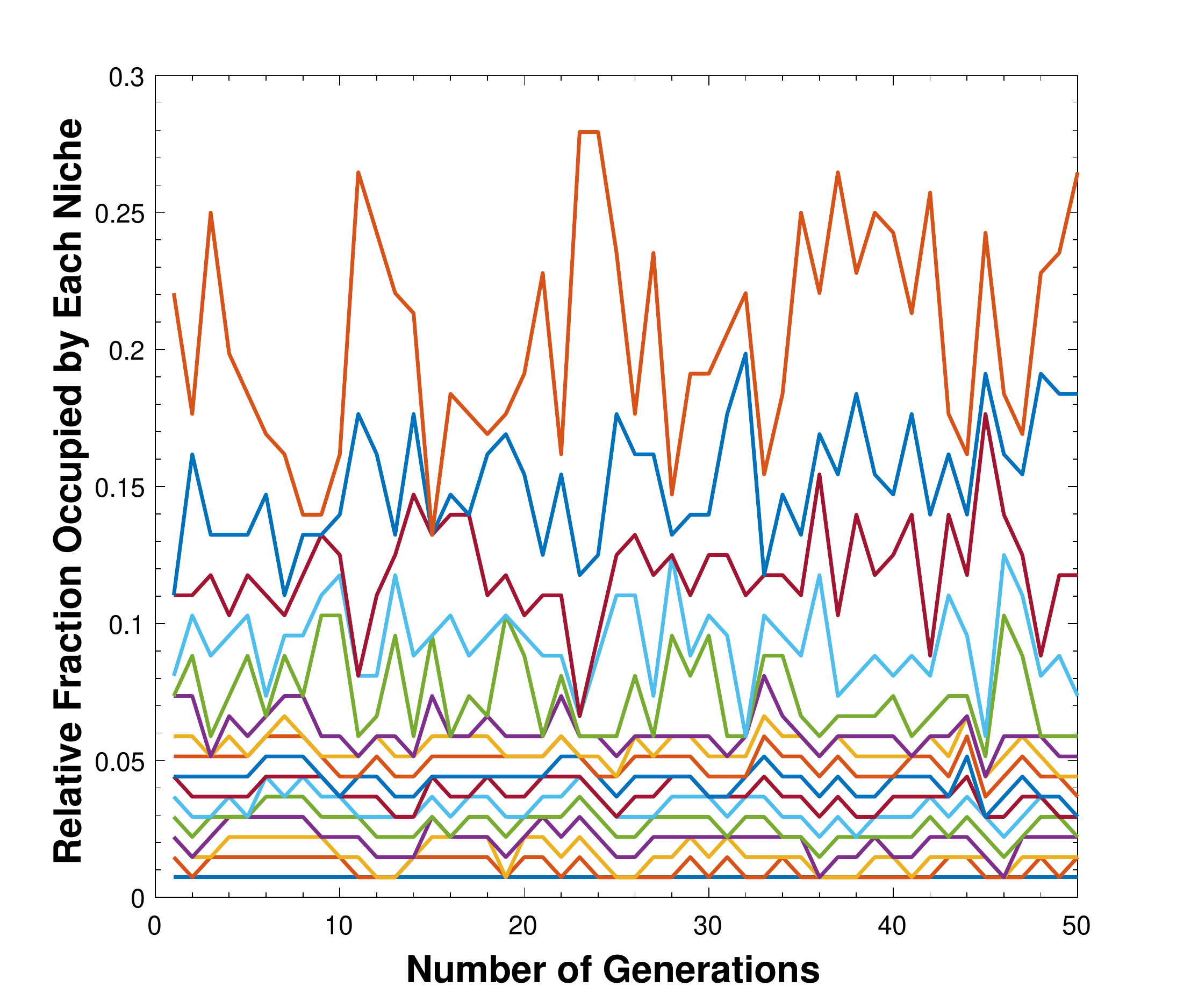}
\captionsetup{justification=centering}
\caption{Case - UGV: Single Stage Variation of Niches}
\label{var_single:Large}
\vspace{-0.5cm}
\end{figure}

Figures \ref{var_multi:Large} and \ref{var_single:Large} show the relative niche sizes as the algorithm proceeds. The largest niche (indicated by the orange line) comprises between $25\% - 35 \%$ of the overall genome population -- which means that no single niche dominates the entire population. This indicates that this niching method preserves diversity. Furthermore, we can see that this niching method is not susceptible to abrupt increases in niche size -- as these abrupt spikes usually settle down within a few generations -- thereby demonstrating the effectiveness of the new niching method. Moreover, it is readily evident that there are other niches of relatively large size, further indicating that different genotypes exist in the final population. However, to answer whether these different genotypes yield intelligence systems capable of exhibiting different behaviors, further investigation is needed in the future.

%Another important observation to consider is that, beside a few neural networks, the best performance of the single stage method is not better than the multi stage method.
%Although different genotypes survived the evolution it is not possible to say if these different genotypes, produce diverse behaviors or not.
\begin{figure}[h]
\centering
\includegraphics[width=0.45\textwidth]{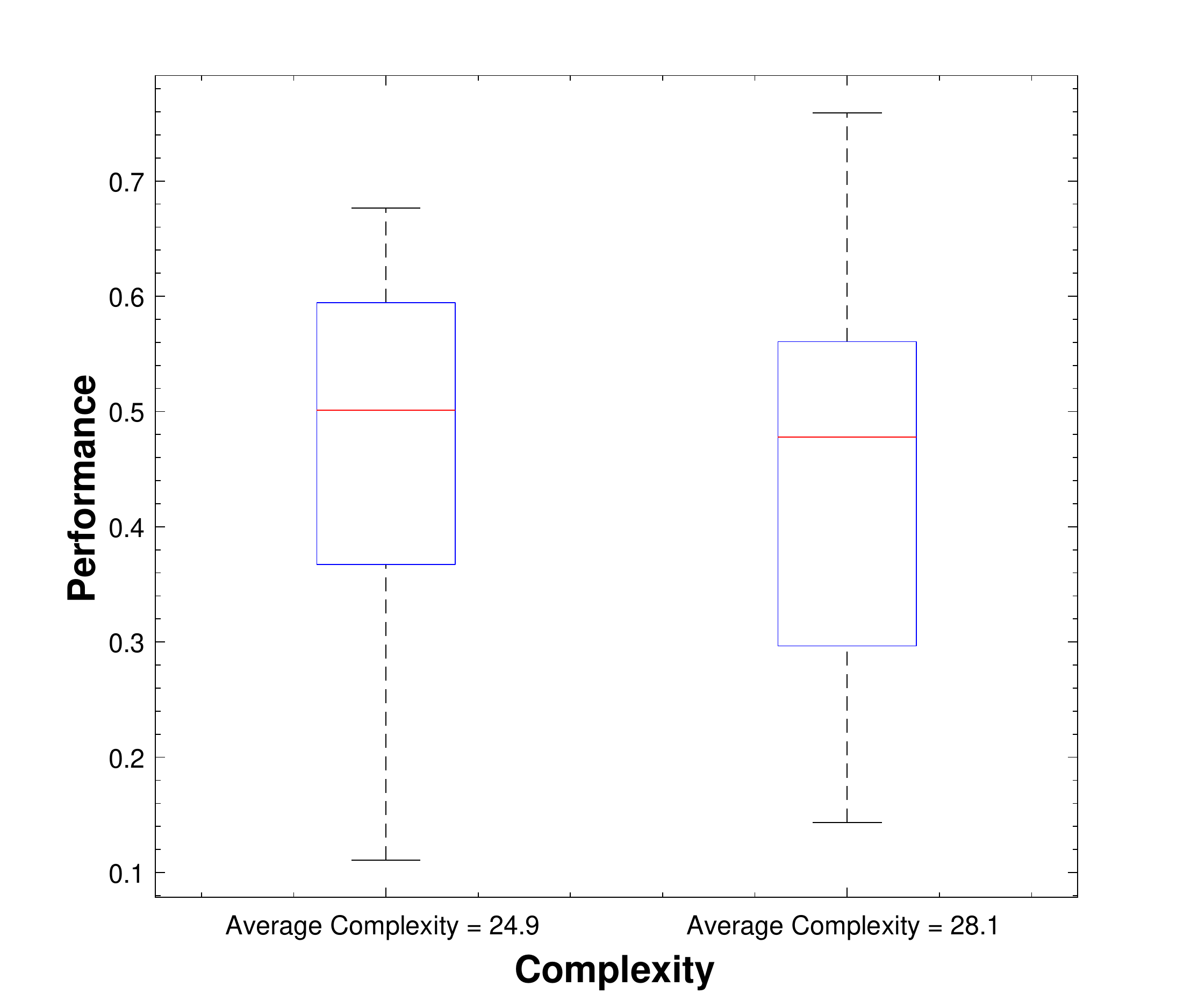}
\captionsetup{justification=centering}
\caption{Case - UGV: Dual Stage Performance Variation w.r.t Complexity}
\label{box_multi:Large}
% \vspace{-0.5cm}
\end{figure}

\begin{figure}[h]
\centering
\includegraphics[width=0.5\textwidth]{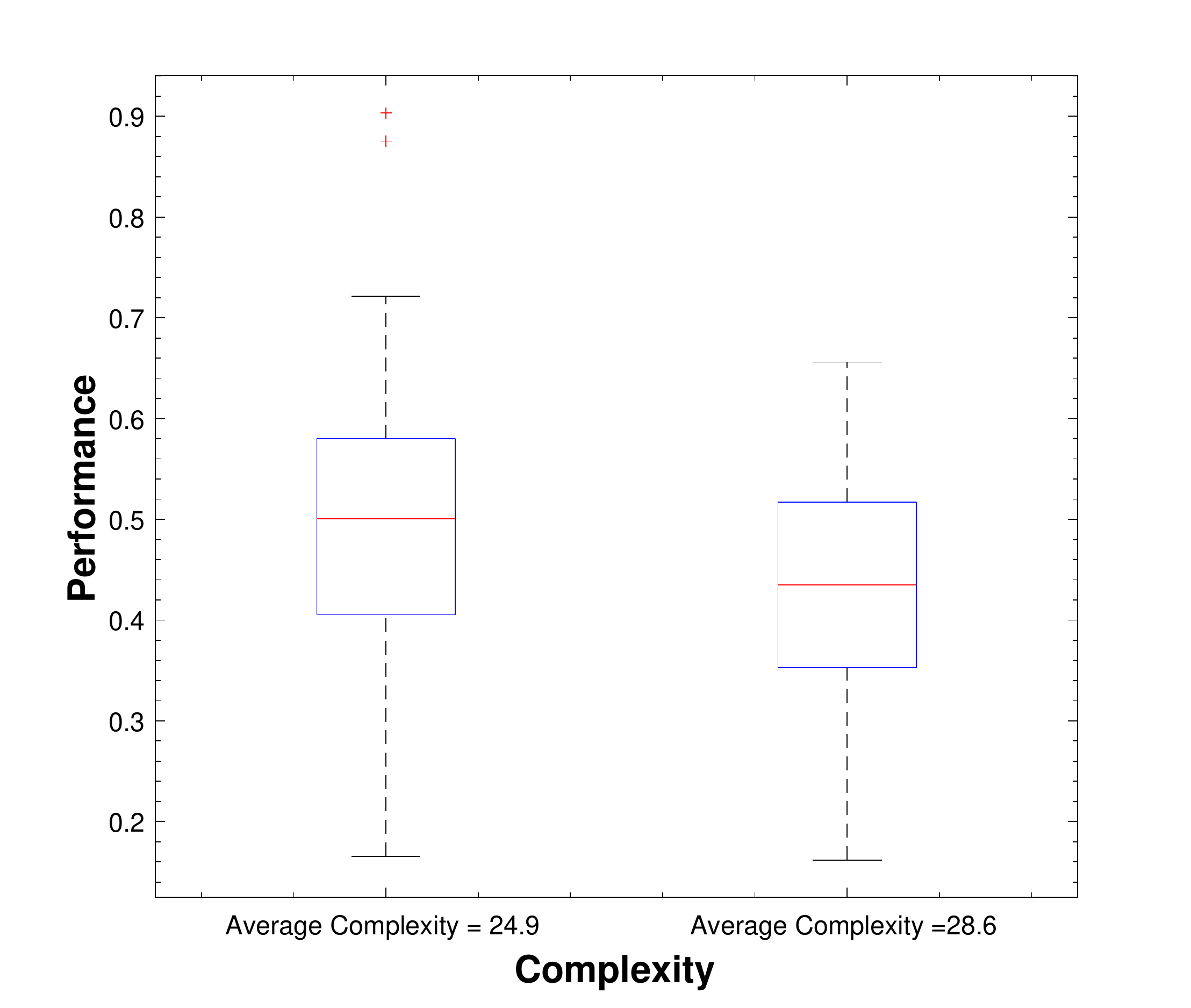}
\captionsetup{justification=centering}
\caption{Case - UGV: Single Stage Performance Variation w.r.t Complexity}
\label{box_single:Large}
\vspace{-0.5cm}
\end{figure}

Figures \ref{box_multi:Large} and \ref{box_single:Large} show the performance variation across ANNs of different complexities from the final population obtained by the single and dual stage optimizations. On average, ANNs of lower complexity provided slightly better performance than ANNs of higher complexity. %It was observed that there was a positive correlation between the performance and complexity for the dual stage optimization; whereas, for the single stage optimization, a negative correlation was seen.
However, the most complex network from the dual stage process outperformed the most complex neural network produced from the single stage process. %The best performance of complex networks using dual stage method is still better than the best complex neural network using a single stage, which shows that dual stage method preserves complexity better than single stage method.
\subsection{Results: Case - Swarm-Robot}
%In this endeavor, the original implementation of multi-objective neuro-evolution was created using MATLAB\textsuperscript{\tiny\textregistered}. Furthermore, the entire case study simulation including modeling the robot kinematics, and sensor behavior were all implemented in MATLAB\textsuperscript{\tiny\textregistered}.
%In order to investigate the effectiveness of the proposed method, two different studies were conducted.
%The first case study compares a two-stage optimization which is proposed before with entire single objective optimization based on the performance. This study aims to investigate the effect of adding multi objective optimization as a start to optimization.
%The second study compares the single objective and multi objective implementations. Based on this study it is possible to compare the effect of the single objective optimization versus multi objective optimization. Because the multi objective only consists of the initial part of the two stage optimization procedure, this study contains fewer generations.
In a similar manner to Case - UGV, two different neuro-evolution processes are undertaken. Here, the networks are trained on a smaller set of $20$ scenarios.
\begin{figure}[h]
\centering
\includegraphics[width=0.45\textwidth]{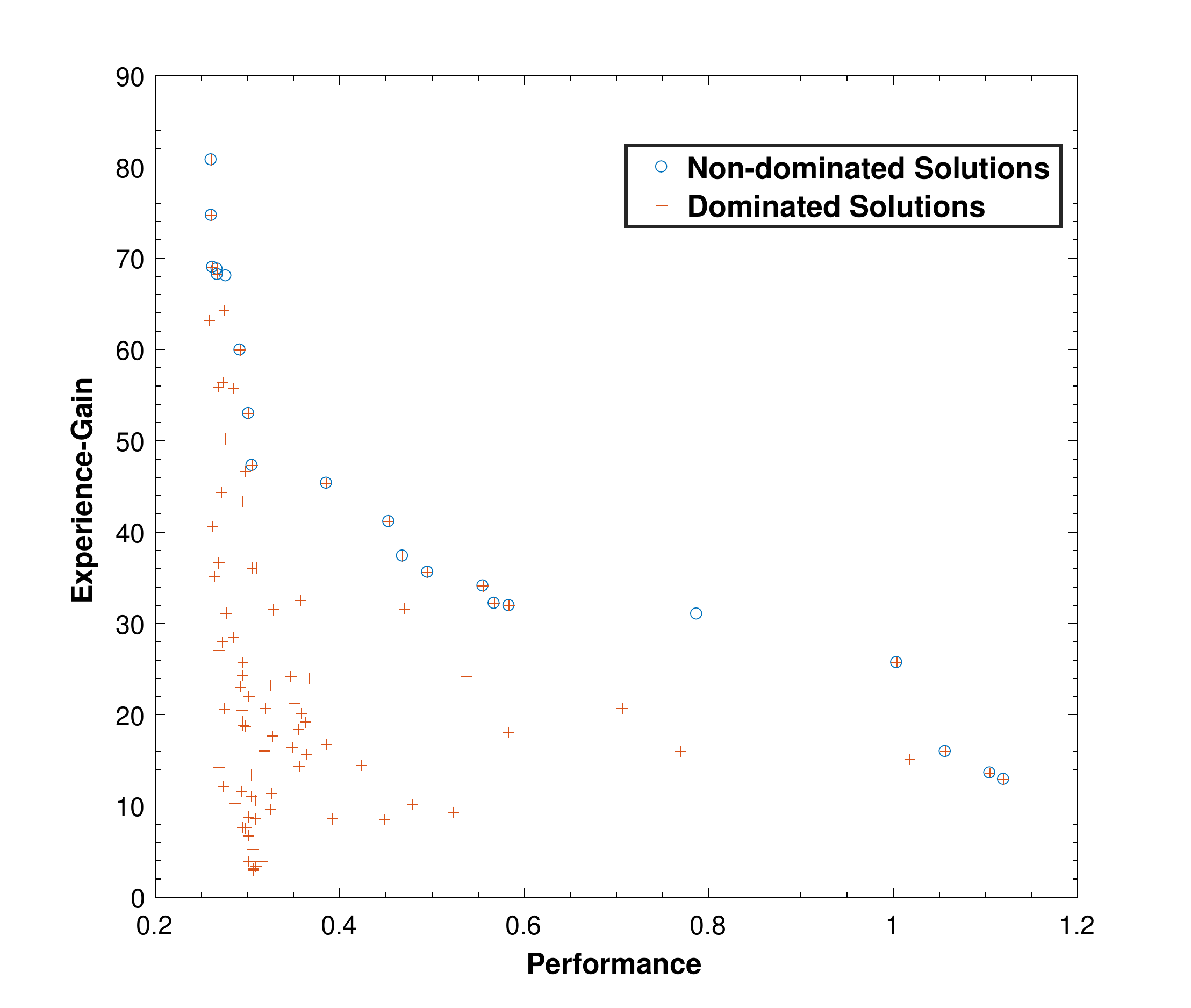}
\captionsetup{justification=centering}
\caption{Case - Swarm-Robot: Pareto Front of the Multi-Objective Optimization Process for Dual stage}
\label{fig:X1}
% \vspace{-0.5cm}
\end{figure}

Figure \ref{fig:X1} shows the top-ranked and the Pareto solutions obtained by the multi-objective optimization portion of the dual stage framework. Note that for the Swarm-Robot, the experience-gain is lower, which is expected owing to the smaller dimensionality of the Experience Space of this robot (fewer sensors). More importantly, significant trade-offs between performance and experience gain are observed for this case as well.

\begin{figure}[h]
\centering
\includegraphics[width=0.5\textwidth]{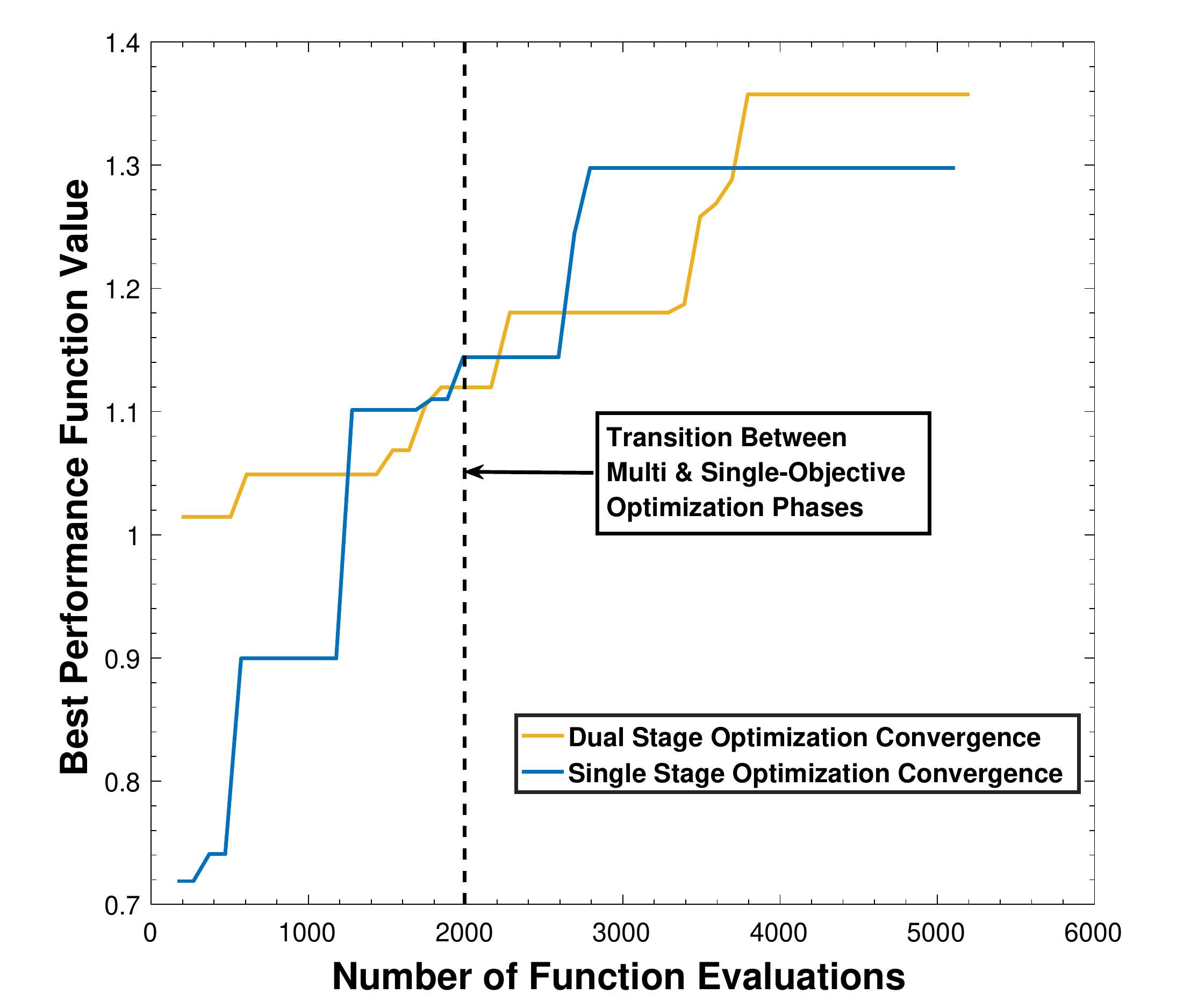}
\captionsetup{justification=centering}
\caption{Case - Swarm-Robot: Comparison of Convergence Histories}
\label{CONV_small}
\vspace{-0.5cm}
\end{figure}

Figure \ref{CONV_small} shows the convergence histories of the dual-stage and single stage optimization processes. Unlike the previous case, with the Swarm-robot, the dual stage process outperforms the single stage process. However, both methods got terminated by the maximum allowed iterations criteria (for the sake of time savings), and hence it is premature to comment on which method could perform better if both are allowed to run till convergence. %It is hard to conclude which method will have better performance at final stage, but it seems that the dual stage method has better performance with a fewer number of function evaluation and can have a comparable performance overall.

\begin{figure}[h]
\centering
\includegraphics[width=0.5\textwidth]{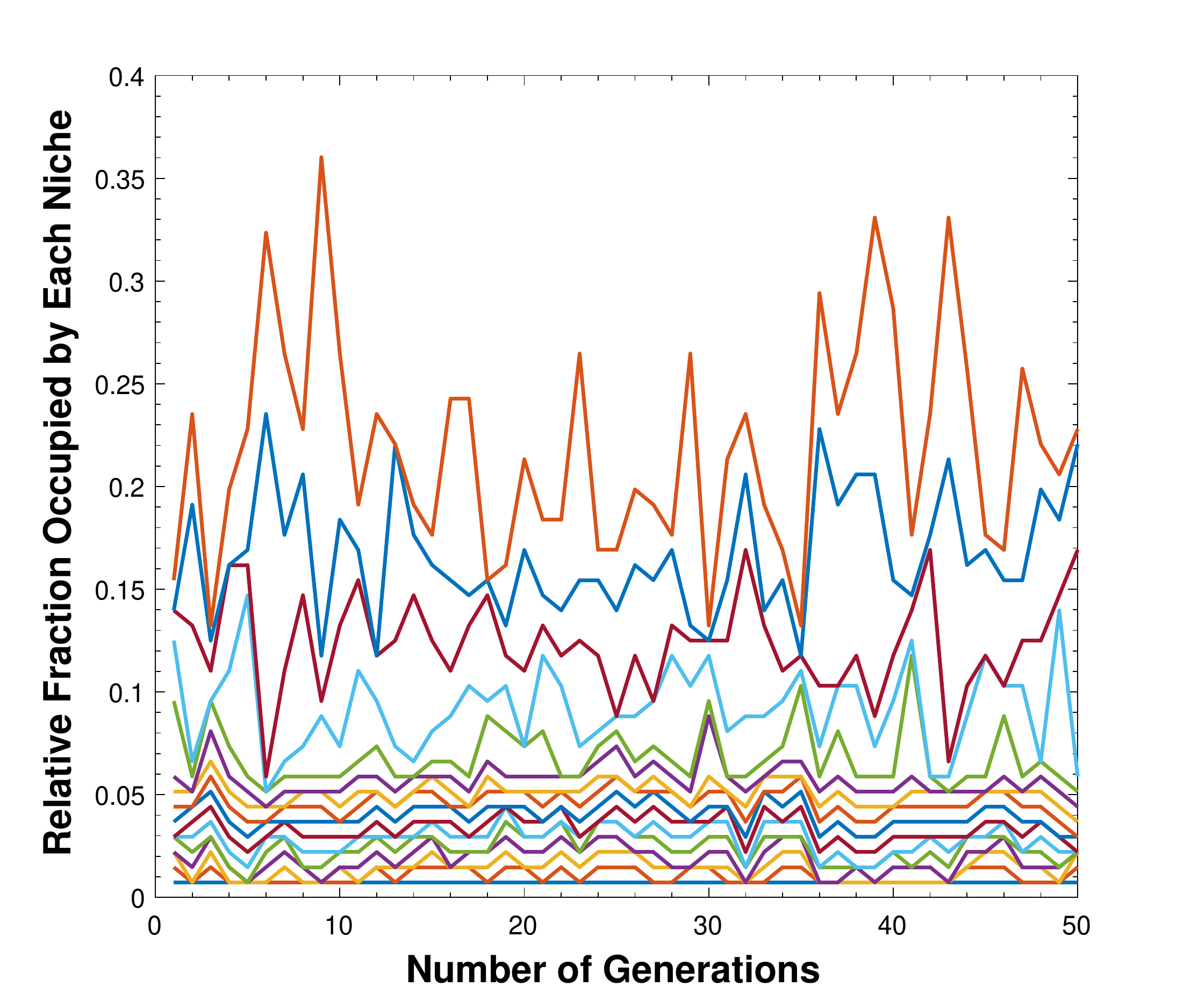}%{NMdfig.pdf}
\captionsetup{justification=centering}
\caption{Case - Swarm-Robot: Dual Stage Variation of Niches}
\label{NicheVar_Multi_Small}
\end{figure}
\vspace{0.5cm}

\begin{figure}[h]
\centering
\includegraphics[width=0.5\textwidth]{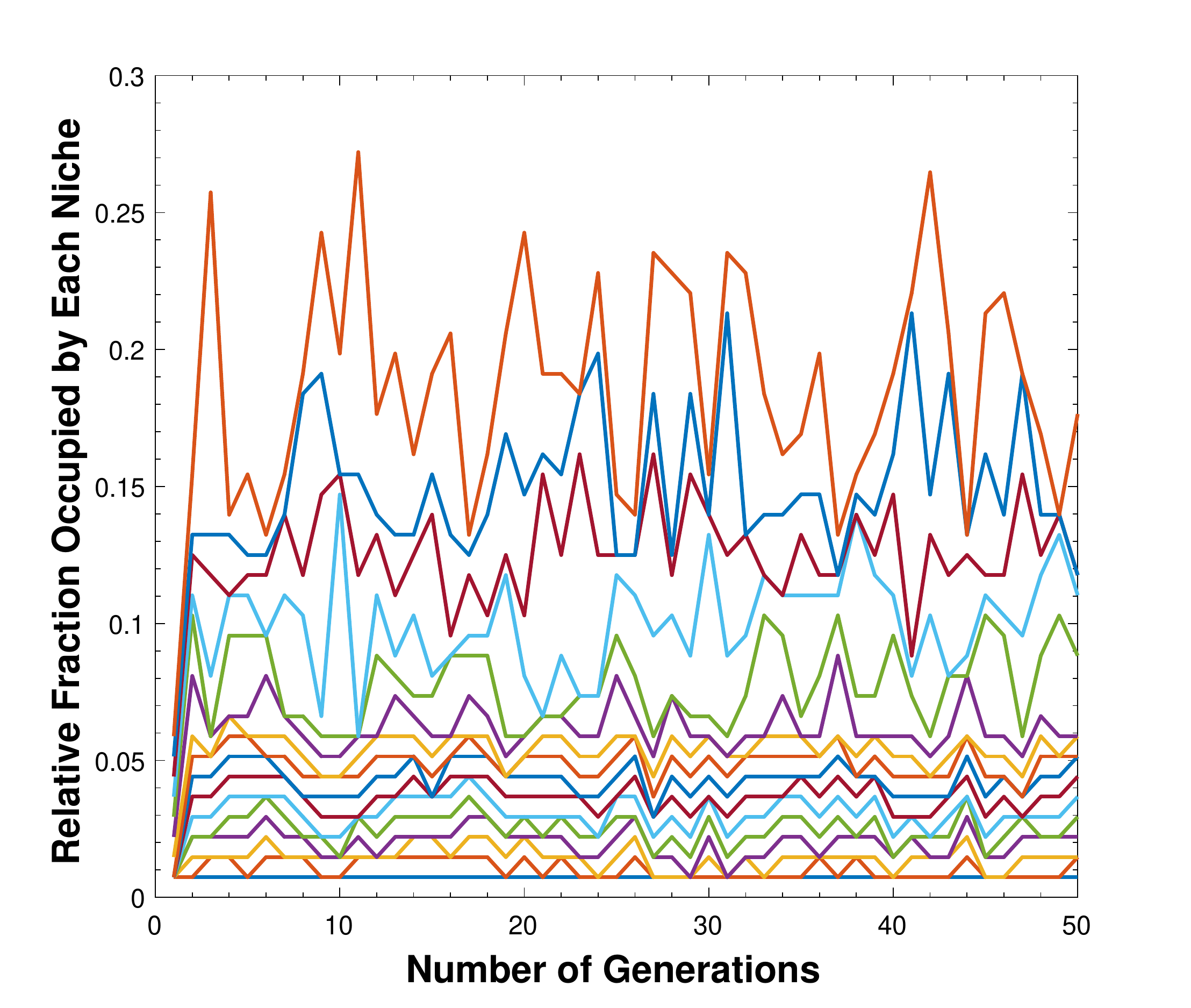}
\captionsetup{justification=centering}
\caption{Case - Swarm-Robot: Single Stage Variation of Niches}
\label{NicheVar_Single_Small}
\end{figure}

Figures \ref{NicheVar_Multi_Small} and \ref{NicheVar_Single_Small} show the size of the niches for the single and dual stage optimizations. From these figures it can be observed that, during both the single stage and dual-stage training processes, our niching methodology prevented any single niche or species from dominating the entire genome population -- the largest niche only consisted of not more than $25 \%$ of the entire population. Although no single niche is dominant, there are a few large niches, attributed by the dominant beneficial qualities of the genotype of those species of ANNs. %Although these behaviors produce good results, they are sufficiently different enough to create different niches, which means different genotypes survived during the evolution process.
\begin{figure}[h]
\centering
\includegraphics[width=0.45\textwidth]{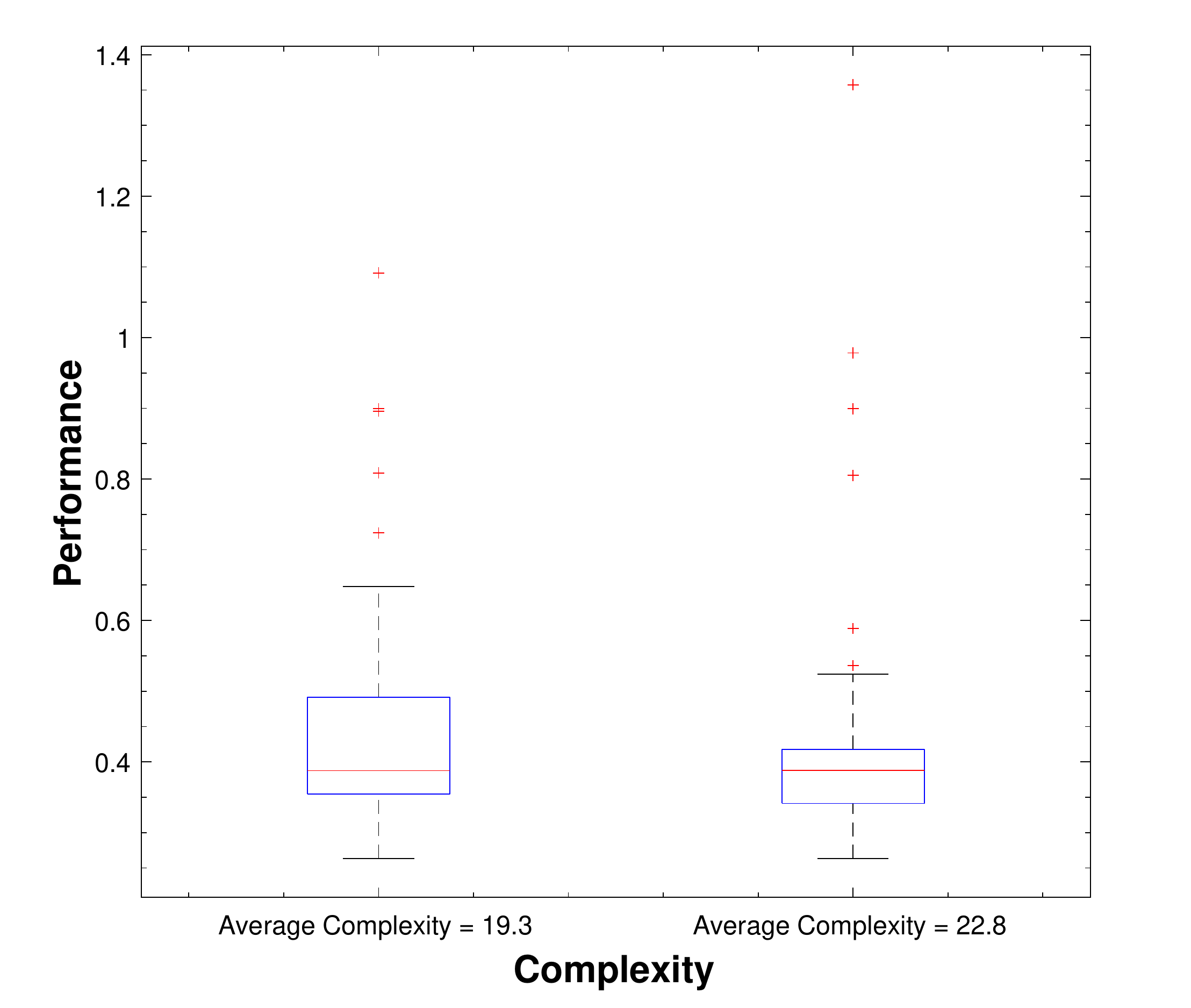}
\captionsetup{justification=centering}
\caption{Case - Swarm-Robot: Dual Stage Performance Variation w.r.t Network Complexity}
\label{box_multi}
% \vspace{-0.5cm}
\end{figure}
\begin{figure}[h]
\centering
\includegraphics[width=0.45\textwidth]{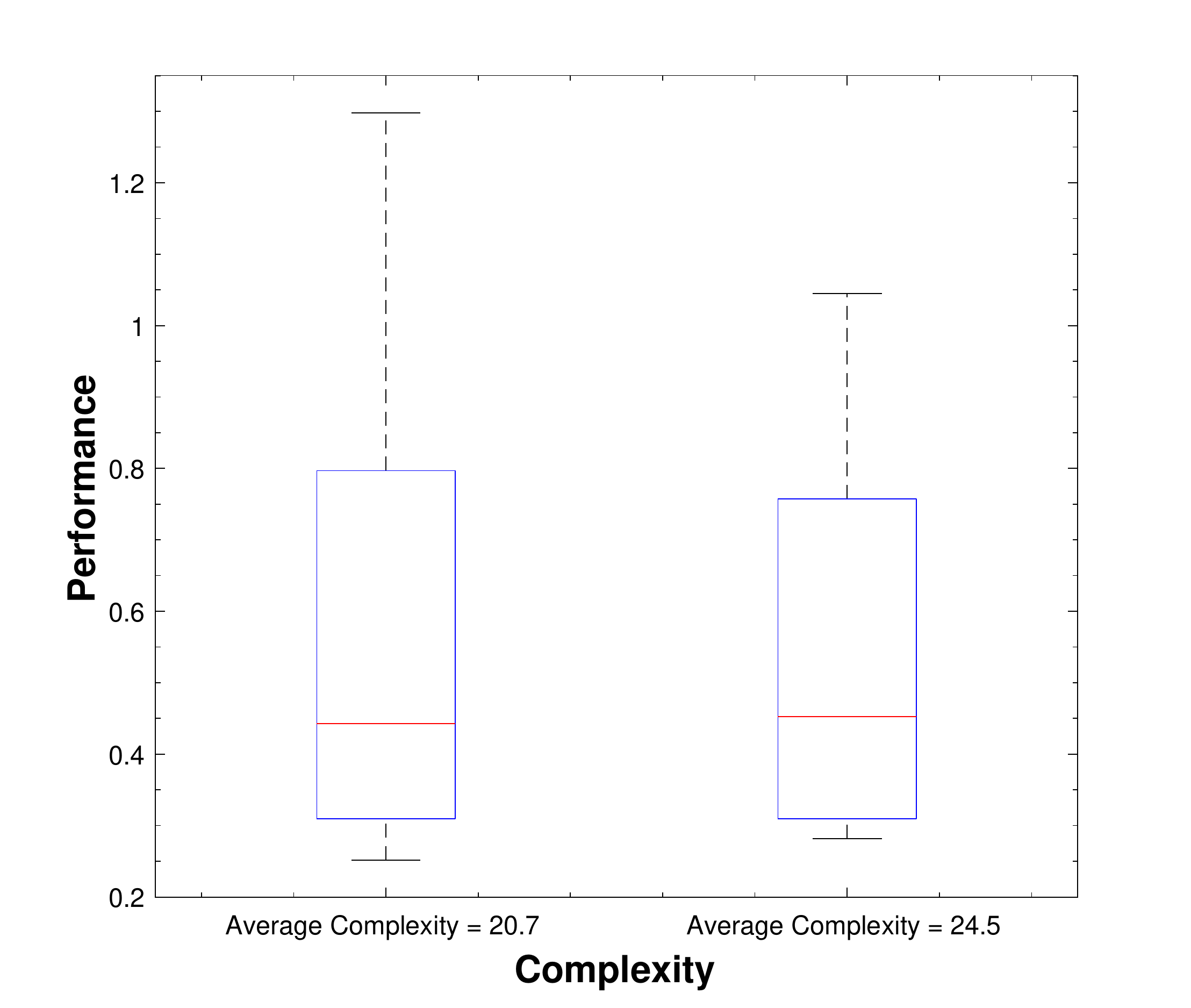}
\captionsetup{justification=centering}
\caption{Case - Swarm-Robot: Single Stage Performance Variation w.r.t Network Complexity}
\label{box_single}
\vspace{-0.5cm}
\end{figure}

Figures \ref{box_multi} and \ref{box_single} show the effects of network complexity on its performance. It is interesting to observe that aside from an anomalous case, there seems to be a negative correlation between performance and complexity in networks produced via the dual stage method. More importantly, the less complex networks appear to exhibit a higher variance in performance compared to the more complex networks, both in the cases of ANNs resulting from dual and single stage processes.
% \begin{figure}[h]
% \centering
% \includegraphics[width=0.5\textwidth]{complex_Perf_Pareto_0.pdf}
% \captionsetup{justification=centering}
% \caption{Case - II: Variation of Network Complexity w.r.t Performance Among Non-Dominated Solutions}
% \label{comp_vs_perf}
% \end{figure}

% Figure \ref{comp_vs_perf} shows the distribution of complexity versus performance among the non-dominated solutions during the final step of the multi objective process. Again, no discernible pattern can be deduced,  which means that perhaps adding complexity may increase the performance in a long time, its short-term effects are minimal.

Owing to the smaller dimensionality of the input space, the Swarm-Robot is expected to have a more difficult time solving the same problem as it gains less information from the environment. Despite this constraint, it is observed that the network converged to a higher objective function value than in Case - UGV. A possible explanation for this could be that, although more inputs can help the system make optimal decisions, training this system might require a greater investment in terms of some generations of evolution and/or population size. %Finally, it could be that, for this problem, an input space of $11$ (as was the case in Case - UGV), might not have been needed.% Finally, the environment for the smaller robot is much smaller, and its speed is also lower than the larger robot. Therefore the probability of collision drastically decreases for the smaller robot.

\subsection{Results: Testing on Unseen Environments}

% \section{MENTOR: TESTING ON UNSEEN ENVIRONMENTS}
% \label{sec:SimExp}
% \subsection{Results: Case - UGV}
For applications where robots will use the ANN-based decision-support to operate in unstructured environment (not necessarily known to the designer), it is critical to measure the performance of the ANN in environments that they have not seen during their training process -- by doing so, we can assess the capability of the intelligence system on combating overfitting.% It is especially important to measure their performance on these environments these systems must be able to handle uncertainties in the operating environment, to an extent. The performance on the trained example might be biased due to overfitting, and therefore the performance on the new environment is a more robust method of assessing the performance.
%
%In an effort to more concretely test our methodology, we sought to implement the aforementioned path planning test problem on a completely new environment and a new robot designed via a high-fidelity simulation platform. More specifically, through this validation effort, we wished to serve two purposes:
% \begin{enumerate}
% \item To justify the utility of training a neural network using MATLAB\textsuperscript{\tiny\textregistered} as opposed to training the system using a high-fidelity software.
% \item To provide an opportunity to test the evolved network on a completely untested environment. This aids in investigating the performance of the decision making system in a new environment.
% \item Investigate how can the designed algorithm work on another robot, without changing the underlying algorithm to understand how much is the algorithm useful for different cases.
% \end{enumerate}
%
To this end, the evolved ANNs for both robotic systems (considered in this paper) are tested on completely new environments created using the V-REP software. All simulations are continued until one of the following conditions are satisfied: the robot reaches within $1 m$ of its goal; the simulation time exceeds $50 s$, or the robot hits the walls or obstacles.

%A new robot is also used which has fewer inputs. The new robot design has only $3$ ultrasonics, and therefore the sensory input is less than an original robot that was used for optimization.
%The new robot is designed based on \underline{REF TO SWARM BOT}. This new robot is a small robot designed for swarm robot tasks, but it was a splendid choice for out algorithm because we could understand the abilities of the algorithm to deal with fewer input.
%The robot size is near $5cm \times 5cm\times 5cm$ and its ultrasonic sensors have the range up to $60 cm$.
The test environment (shown in Fig. \ref{VREP_PIC}, in the Appendix) spans an area of dimensions $12m \times 12m$ and $4 m\times 4 m$ for Case - UGV and Case - Swarm-Robot, respectively. Both the size and the layout of the obstacles in these new test environments are different from the environments on which the ANNs were trained; however, the distance between the robot's starting position and destination are kept within the same range in the training and testing environments to ensure reasonable simulation time.
%The newly designed environment created in V-REP can be seen in the Appendix.
% Fig. \ref{VREP_PIC}.

%The robot I wanted to add about the robot, but I think we should describe the robot before
The performance of the best neural network obtained by the single stage and the dual stage training procedures are tested on the newly created environments, using 10 randomly selected pairs of starting and destination points. For the Case - UGV, both the single and dual stage optimized networks failed to perform well. The single stage network was unable to reach the goal in all examples, while the dual stage network could reach its destination in only one scenario. On the other hand for Case - Swarm-Robot, both the single stage and dual stage optimized networks performed well and successfully reached the destination in multiple scenarios; the dual stage network was capable of reaching the destination faster, but the single stage network provided greater survivability -- thus two different behaviors were exhibited. %It seems that the performance of the neural network evolved in Case - Swarm-Robot was not hampered, even though it was subjected to fewer training scenarios. This could possibly be due to the fact that navigating the Swarm-Robot is easier in this environment, and that there are fewer sensors (Experience Space), thus making the training process less burdensome.

The problem of overfitting appears to be more severe for Case - UGV as it has a larger input space, thus increasing the training burden on the algorithm. Therefore the dual-stage performed better in this case which lends credence to our hypothesis that the dual stage training paradigm can be used when training the system is computationally cumbersome. The dual stage method offers greater opportunities for the candidate population to explore compared to the single stage, thus potentially opening opportunities to find optimal results with limited training. In addition, the dual stage method promotes behavioral diversity early on in the neuro-evolution process. However, care must be exercised when transitioning to the single-objective phase, so as to not skew the total population towards the experience-gain objective rather than the performance objective.

\section{CONCLUSION}\label{sec:conclusion}
In this paper, we propose a method, called MENTOR, for evolving neural network topologies with rank-based elitist non-dominated sorting, thereby allowing the consideration of multiple desired criteria. In addition to enabling the capacity of multi-objective search, important advancements made to the neuro-evolution of augmenting topologies paradigm include the incorporation of a new niching technique to preserve speciation and modification of the mutation operator. Robot path planning problems are formulated, and MENTOR is used to train ANNs that serve as optimal navigational decision-support in environments where the robot has to reach a target destination while avoiding randomly placed/randomly sized obstacles. In order to prevent overfitting and avoid premature convergence, the idea of goal-cognizant experience-gain criteria is explored. Thus, instead of solely optimizing performance, a multi-objective optimization of performance and experience gain is conducted. The experience is defined in a way that helps the system to preserve appropriate performance levels while assisting the optimization to diversify the behaviors being encoded by the ANN. To further explore this premise, a two stage optimization is proposed -- in the first stage, multi-objective neuro-evolution is performed which leads to a diverse population that is then used to initiate the single objective neuro-evolution stage. The performance of the dual stage approach is compared to that accomplished by solely using single objective neuro-evolution.

In the training environment, the ANNs resulting from the single stage method performed slightly better for a small UGV case, while the ANN resulting from the dual stage method performed better for the swarm-robot case. However, in unseen test environments, the ANNs generated by the dual stage method performed better or as good as that resulting from the single stage method. These observations call for further investigation of the topological variations and associated rate of weight stabilization occurring in single- vs. multi-objective neuro-evolution -- which is an immediate next step in this research. Future work will also explore approaches to better exploit the experience gain through intra-generational learning, perform benchmark testing on problems from the Open AI gym domain, and explore opportunities to evolve deep ANN topologies. Finally, the physical implementation of ANN-based decision-support models constructed by MENTOR on practical autonomous systems will allow more critical evaluation of the MENTOR training process. %It appears that the initial multi-objective optimization procedure caused the diversity of the population to decrease, thereby causing the fitness in the subsequent stage to stagnate.
%Talk about limited options in testing MONE
%\appendix
%\begin{acks}
%\end{acks}
%Talk about limited options in testing MONE
%\appendix
%\begin{acks}
%\end{acks}

% \vspace{-0.5cm}
\bibliographystyle{IEEEtran}
\bibliography{bibliography}

\vspace{1 cm}

% \newpage
\appendix
% \section*{APPENDIX}

%\section{Algorithm for Experience-Gain}

\vspace{-1cm}
\begin{figure}[H]
\centering
\includegraphics[width=0.4\textwidth]{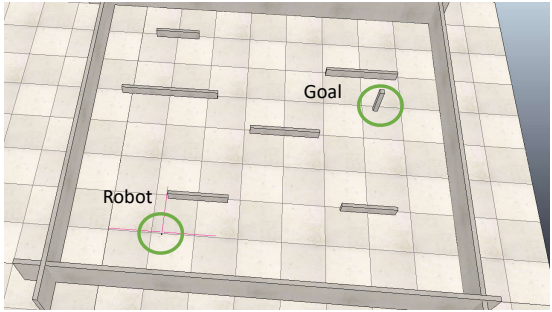}
\captionsetup{justification=centering}
% \vspace{-1.0cm}
\caption{Test/Unseen Environment Created in V-REP for Case - Swarm-Robot}
\label{VREP_PIC}
\end{figure}

% \begin{algorithm}
% \SetAlgoNoLine
% {
% $C(n)=n$ (tag connected components for all nodes)

% Sort All Edges based on the weights, Ascending

% $m=0$,
% $i=0$,$MST= empty$

% \While {$m \leq n-1$}
% {
% 	$i++$

%     \If { $C(e_{i,s})!=C( e_{i,f})$  } {
%     %(two nodes of $e_{i}$ are from different components)
%      Add $e_{i}$ to  $MST$
%       combine $C(e_{i,s}), C( e_{i,f})$
%     }
% }
% $Return (MST)$
% }
% \caption{Kruskal Algorithm}
% \label{Kruskal}
% \end{algorithm}

\begin{algorithm}
 \caption{Kruskal Algorithm}
 \begin{algorithmic}[1]
 \renewcommand{\algorithmicrequire}{\textbf{Input:}}
 \renewcommand{\algorithmicensure}{\textbf{Output:}}
 \REQUIRE Graph
 \ENSURE  Minimum Spanning Tree

$C(n)=n$ (tag connected components for all nodes)

Sort All Edges based on the weights, Ascending

$m=0$,
$i=0$,$MST= empty$

\While {$m \leq n-1$}
{
	$i++$

    \If { $C(e_{i,s})!=C( e_{i,f})$  } {
    %(two nodes of $e_{i}$ are from different components)
     Add $e_{i}$ to  $MST$
      combine $C(e_{i,s}), C( e_{i,f})$
    }
}
$Return (MST)$
 \end{algorithmic}
 \end{algorithm}

% \bibliographystyle{IEEEtran}
% \bibliography{payam2018map}

\end{document}